%% file: acl_latex.tex
\documentclass[11pt]{article}

\usepackage[final]{acl}

\usepackage{times}
\usepackage{latexsym}
\usepackage{tcolorbox}
\usepackage{multirow}
\tcbuselibrary{skins} 
\newtcolorbox{diagnosisbox}{
    enhanced,
    colback=black!5,      
    colframe=black!75,    
    boxrule=0.5pt,        
    sharp corners,        
    left=6pt,             
    right=6pt,
    top=6pt,
    bottom=6pt
}
\usepackage[T1]{fontenc}

\usepackage[utf8]{inputenc}

\usepackage{microtype}

\usepackage{inconsolata}

\usepackage{graphicx}
\usepackage{hyperref}
\usepackage{url}
\usepackage{amssymb}
\usepackage[table]{xcolor}
\usepackage{booktabs}
\usepackage{array}
\usepackage{xcolor}
\usepackage{algorithm}
\usepackage{algpseudocode}
\usepackage{wrapfig}
\usepackage{amsmath}
\usepackage{mathtools}
\usepackage{fontawesome5} 
\usepackage{hyperref}     
\usepackage{amsmath}

\definecolor{myred}{RGB}{232, 149, 153}
\definecolor{mypurple}{RGB}{168, 168, 202}
\definecolor{mygreen}{RGB}{122, 208, 191}

\title{From Competition to Synergy: Unlocking Reinforcement Learning for Subject-Driven Image Generation}
\author{
 \textbf{Ziwei Huang\textsuperscript{1}},
 \textbf{Yin Shu\textsuperscript{2}},
 \textbf{Hao Fang\textsuperscript{2}},
 \textbf{Quanyu Long\textsuperscript{3}},
 \textbf{Wenya Wang\textsuperscript{3}},
 \\
 \textbf{Qiushi Guo\textsuperscript{2}},
 \textbf{Tiezheng Ge\textsuperscript{2}},
 \textbf{Leilei Gan\textsuperscript{1}\thanks{Corresponding author}},
\\
\\
 \textsuperscript{1}Zhejiang University,
 \textsuperscript{2}Alibaba Group,
 \textsuperscript{3}Nanyang Technological University,
\\
\texttt{\{ziweihuang, leileigan\}@zju.edu.cn}
}

\begin{document}
\maketitle
\begin{abstract}
Subject-driven image generation models face a fundamental trade-off between identity preservation (fidelity) and prompt adherence (editability). While online reinforcement learning (RL), specifically GRPO, offers a promising solution, we find that a naive application of GRPO leads to competitive degradation, as the simple linear aggregation of rewards with static weights causes conflicting gradient signals and a misalignment with the temporal dynamics of the diffusion process.
To overcome these limitations, we propose Customized-GRPO, a novel framework featuring two key innovations: (i) Synergy-Aware Reward Shaping (SARS), a non-linear mechanism that explicitly penalizes conflicted reward signals and amplifies synergistic ones, providing a sharper and more decisive gradient. (ii) Time-Aware Dynamic Weighting (TDW), which aligns the optimization pressure with the model's temporal dynamics by prioritizing prompt-following in the early, identity preservation in the later. Extensive experiments demonstrate that our method significantly outperforms naive GRPO baselines, successfully mitigating competitive degradation. Our model achieves a superior balance, generating images that both preserve key identity features and accurately adhere to complex textual prompts.\footnote{\faGithub~\href{https://github.com/Safeoffellow/Customized-GRPO}{github.com/Safeoffellow/Customized-GRPO}}
\end{abstract}

\input{section/introduction}
\input{section/related_work}
\input{section/preliminary}
\input{section/pilot}
\input{section/method}
\input{section/experiments}
\bibliography{custom}

\clearpage

\appendix
\input{section/appendix}

\end{document}

%% file: section/introduction.tex
\section{Introduction}
\label{intro}
\begin{figure}[ht]
    \centering
    \setlength{\abovecaptionskip}{0.2cm}
    \setlength{\belowcaptionskip}{-0.3cm}
    \includegraphics[width=0.42\textwidth]{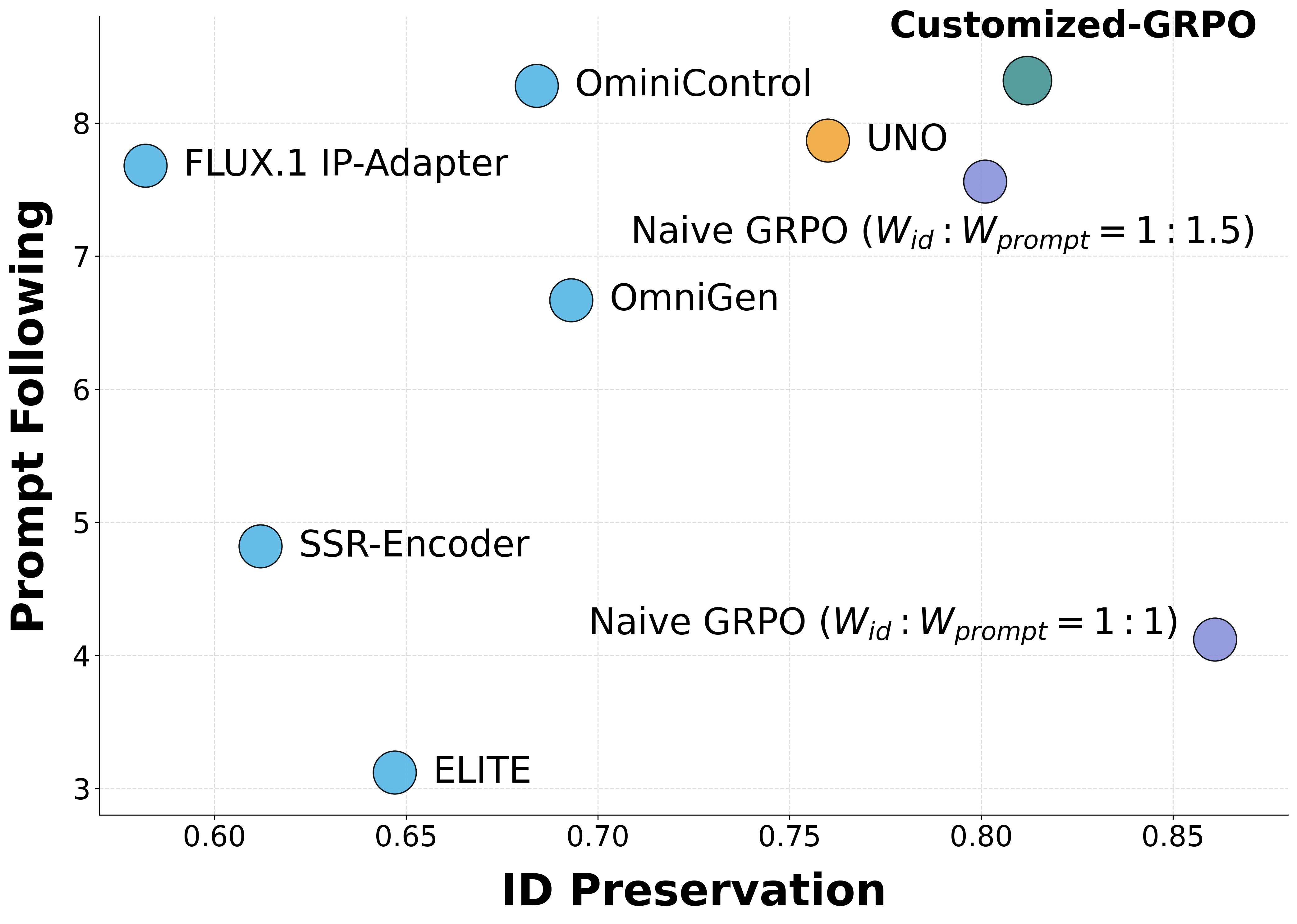}
    \caption{\textbf{Customized-GRPO} achieves a state-of-the-art balance between ID Preservation and Prompt Following on DreamBench. Our method and Naive GRPO are built upon UNO.}
    \label{fig:dot}
    \vspace{-3mm}
\end{figure}

\begin{figure*}[htbp]
    \centering
    \setlength{\abovecaptionskip}{0.2cm}
    \setlength{\belowcaptionskip}{-0.3cm}
    \includegraphics[width=0.75\textwidth]{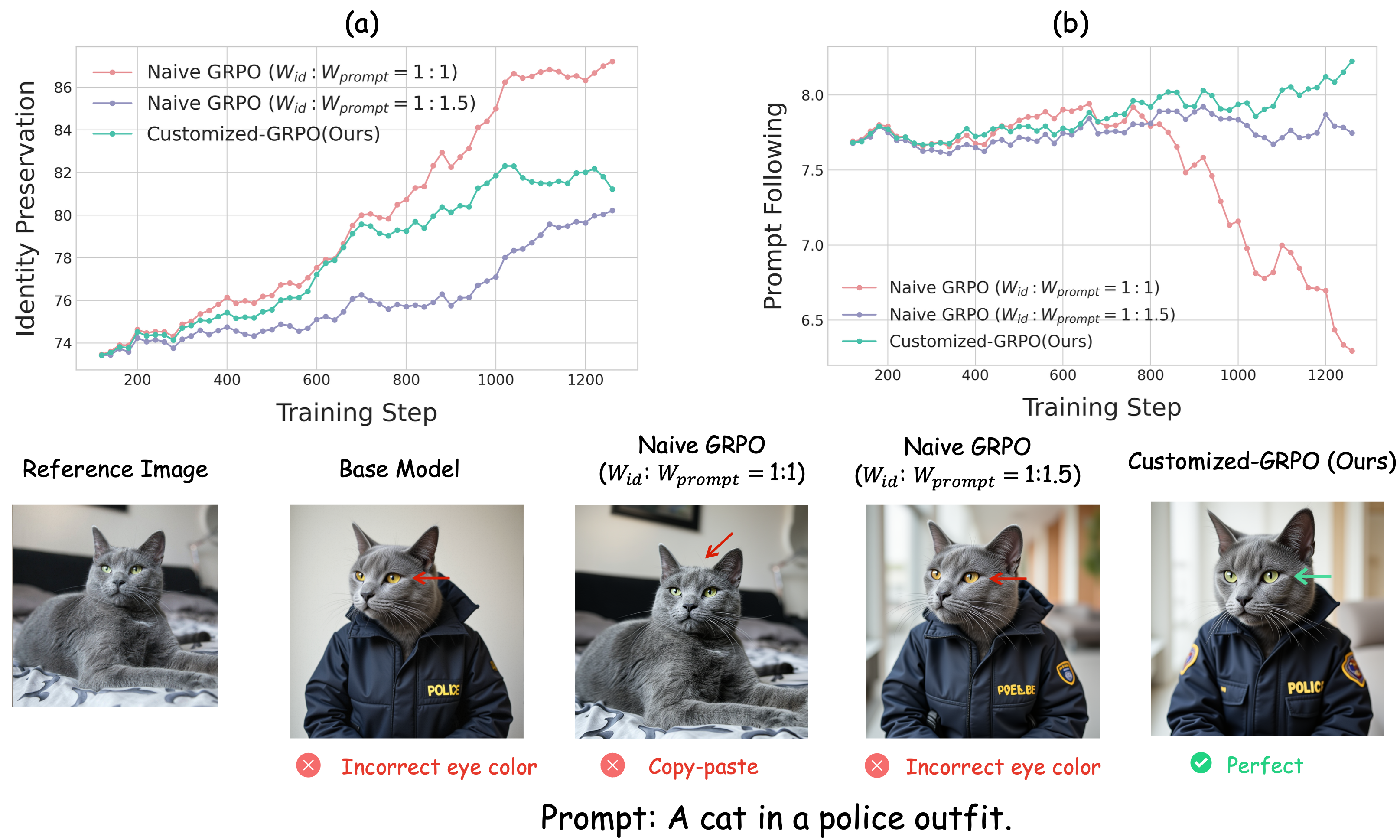}
    \caption{\textbf{Comparison of Naive GRPO with fixed linear weights against our Customized-GRPO}. The (a, b) training curves show that Naive GRPO with 1:1 weights (\textcolor{myred}{red}) overfits to identity, causing a collapse in prompt following while 1:1.5 weights (\textcolor{mypurple}{purple}) prevents this collapse at the cost of poor identity learning. Our method (\textcolor{mygreen}{green}) improves both capabilities concurrently. The qualitative results (bottom) confirm these findings.}
    \label{fig:hps_train}
    \vspace{-3mm}
\end{figure*}

In recent years, subject driven image generation, which aims to create customized images that align with both a textual prompt and the specific subjects in reference images, has garnered substantial interest across both academic and industrial communities~\citep{blip_diffusion, dreambooth, textual_inversion, ipadapter, ssr_encoder, flexip, uno}. 
Unlike standard text-to-image (T2I) generation \citep{stablediffusion, sdxl, mars}, which focuses solely on text–image alignment, this task demands a dual capability: maintaining high-fidelity identity preservation from reference images and ensuring accurate prompt adherence in novel contexts.
This dual objective introduces a critical generalization challenge: the model must learn a subject representation that is robust enough to preserve its core identity, yet adaptive enough to integrate seamlessly into new contexts as indicated in the prompt. Achieving this balance between identity fidelity and prompt adherence is central to realizing practical personalized generation.

To address this challenge, current research has primarily evolved along two main paradigms: finetuning-based and adapter-based customization.
Finetuning-based approaches, such as DreamBooth~\citep{dreambooth} and Textual Inversion~\citep{textual_inversion}, enable subject-specific customization by fine-tuning model parameters using a small set of reference images to learn the specialized representation of single subject. 
However, these methods are computationally intensive and requires retraining for each new subject, making it impractical for scalable customization.
Instead, Adapter-based approaches~\citep{uno,ipadapter,ominicontrol,omnigen,syncd} introduce lightweight cross-modal adapters to bridge text and visual representations without requiring per-subject fine-tuning. 
However, in practice, they often exhibit poor generalization when confronted with complex prompts or novel subjects.
To mitigate this, recent methods have explored more generalized strategies, ranging from large-scaled pre-training~\citep{syncd} to sophisticated attention calibration and network designs that attempt to disentangle identity and prompt representations~\citep{latentunfold, conceptrol, flexip}.
Nonetheless, essentially, all of the above methods rely on Supervised Fine-Tuning (SFT), which operates as a form of behavior-cloning paradigm that  lacks the capacity to generalize beyond the observed examples. 
This limitation becomes more pronounced when the model encounters novel prompts or unseen subjects, where the training distribution no longer provides reliable guidance. 

To move beyond the behavior-cloning by supervised learning, we adopt the paradigm of online Reinforcement Learning (RL).
Recent advances, Group Relative Policy Optimization (GRPO) \citep{dancegrpo, flowgrpo, ar_grpo}, have shown that directly optimizing generation policies against learned reward functions can significantly enhance text-to-image (T2I) alignment and perceptual quality.
However, extending GRPO to the subject-driven domain introduces new complexity. 
Unlike general T2I settings that pursue a single goal such as text–image consistency, subject-driven generation task requires multi-objective alignment between identity fidelity and prompt adherence, two competing goals as illustrated in Figure \ref{fig:hps_train}. We find that naive GRPO with fixed linear weighting fails to reconcile these competing rewards: heavier identity weighting causes the model to overfit the subject and ignore the prompt, while increasing prompt weight restores editability at the cost of fidelity.
This imbalance arises because linearly aggregated rewards produce weak and ambiguous gradients, while static weighting neglects the coarse-to-fine dynamics of diffusion.
Consequently, the optimization oscillates between objectives and result in competitive degradation, where improving one capability inevitably degrades the other.


In this work, to overcome competitive degradation, we introduce \textbf{Customized-GRPO}, a novel approach featuring two synergistic innovations. First, to address the issue of reward conflict, we introduce Synergy-Aware Reward Shaping (SARS), a Pareto-inspired mechanism that provides a sharp and decisive learning signal by explicitly penalizing misaligned advantage signals and rewarding synergistic ones. Second, to tackle the problem of static optimization pressure, we develop Time-Aware Dynamic Weighting (TDW). Motivated by a Fourier analysis of the denoising process, this method aligns the optimization objective with the model's coarse-to-fine generation trajectory by dynamically allocating weights to each reward based on the current timestep.

Through extensive experiments, we demonstrate that Customized-GRPO successfully achieves a superior balance and generating images that are both faithful to the subject's identity and accurately aligned with complex textual prompts, as validated by the improvements shown in Figure \ref{fig:dot} and \ref{fig:hps_train}.

%% file: section/related_work.tex
\section{Related Work}
\label{related work}
\subsection{Subject-Driven Generation}

Subject-driven image generation aims to synthesize novel images of a specific concept provided through a few reference images, while adhering to the guidance of a textual prompt. Early finetuning-based methods like DreamBooth~\citep{dreambooth} and Textual Inversion~\citep{textual_inversion} achieve per-subject customization through fine-tuning, but this process is computationally expensive and must be repeated for each new concept.

To overcome this limitation, a significant body of work has focused on adapter-based methods. A popular approach involves training lightweight adapters to inject visual conditions into the model's attention layers~\citep{ipadapter, ominicontrol, uno, flexip, ssr_encoder}. More recently, a distinct inference-time paradigm has emerged. These methods~\citep{diptych, latentunfold, conceptrol} require no additional training and typically operate by reframing the task as a form of guided inpainting or by directly manipulating the model's attention maps to mitigate concept confusion.

\subsection{GRPO in Text-to-Image Generation}
Reinforcement Learning (RL), particularly methods designed for aligning Large Language Models (LLMs) with human feedback, has recently emerged as a powerful paradigm for enhancing visual generation models. Among these, Group Relative Policy Optimization (GRPO)~\citep{deepseek} has become prominent due to its training stability and efficiency.

\vspace{-1mm}
\paragraph{GRPO in Autoregressive Models.} The application of GRPO to autoregressive (AR) visual models is a natural extension of its success in the text domain. AR-GRPO~\citep{ar_grpo} adapt the framework to fine-tune AR image generators for better alignment with human perceptual preferences. Subsequent research has leveraged GRPO to unlock more complex capabilities. T2I-R1~\citep{t2i_r1} and GoT-R1~\citep{got_r1} introduce Chain-of-Thought-inspired planning stages, using GRPO with sophisticated reward systems to guide the model towards discovering superior semantic-spatial reasoning strategies for complex compositional generation.

\vspace{-1mm}
\paragraph{GRPO in Diffusion and Flow Matching Models.} Applying online RL to non-autoregressive models like diffusion and flow matching poses a greater challenge, as their standard deterministic sampling processes lack the stochasticity required for exploration. A key breakthrough is the introduction of an \textbf{ODE-to-SDE conversion} during training, which injects controllable noise to enable policy learning.
DanceGRPO~\citep{dancegrpo} and Flow-GRPO~\citep{flowgrpo} are foundational in this area, establishing a stable and scalable framework for applying GRPO across diverse generative paradigms and tasks. Building upon this, MixGRPO~\citep{mixgrpo} further enhance training efficiency by proposing a hybrid mixed ODE-SDE sampling strategy, confining the computationally intensive SDE exploration to a small sliding window of timesteps.

%% file: section/preliminary.tex
\section{Problem Formulation}

\paragraph{Diffusion Model.} A diffusion process gradually destroys an observed datapoint $\mathbf{x}$ over timestep $t$, by mixing data with noise, and the forward process of the diffusion model can be defined as~\citep{denoising}:
\begin{equation}
\label{ddpm_forward}
\mathbf{z}_t=\alpha_t\mathbf{x}+\sigma_t\boldsymbol{\epsilon},\mathrm{~where~}\boldsymbol{\epsilon}\sim\mathcal{N}(0,\mathbf{I}),
\end{equation}
and $\alpha_t$ and $\sigma_t$ denote the noise schedule. The noise schedule is designed in a way such that $\mathbf{z}_{0}$ is close to clean data and $\mathbf{z}_{1}$ is close to Gaussian noise.
To generate a new sample, we initialize the sample $\mathbf{z}_1$ and define the sample equation of the diffusion model given the denoising model output $\hat{\boldsymbol{\epsilon}}$ at time step $t$:
\begin{equation}
\mathbf{z}_s=\alpha_s\hat{\mathbf{x}}+\sigma_s\hat{\boldsymbol{\epsilon}}, 
\label{ddim}
\end{equation}
where $\hat{\mathbf{x}}$ can be derived via Eq.\eqref{ddpm_forward} and then we can reach a lower noise level $s$.

\vspace{-1mm}
\paragraph{Flow Matching.} One drawback of this iterative denoising process is it can be computationally expensive and slow. 
To address this, Flow Matching models~\citep{flow} directly learn the velocity field of the data transformation, enabling faster generation without relying on slow step-by-step denoising.
In the rectified flow~\citep{flow}, a specific form of flow matching, the forward process is defined as a linear interpolation between the true data sample $x_0 \sim X_0$ and $x_1 \sim X_1$ denote a noise sample: 
\begin{equation}
    \mathbf{z}_t = (1-t)\mathbf{x_0} + t\mathbf{x}_1 \quad 
    \label{eq:rectified_flow_forward}
\end{equation}
Then a transformer model is trained to directly predict the velocity field $v_\theta(z_t, t)$ by minimizing the Flow Matching objective:
\begin{equation}
\label{eq:flow_loss}
    \mathcal{L}(\theta) = \mathbb{E}_{t,\, \mathbf{x}_0 \sim X_0,\, \mathbf{x}_1 \sim X_1}
    \left[
        \left\| v - v_\theta(\mathbf{z}_t, t) \right\|^2
    \right]
\end{equation}
where $v$ is the ground-truth velocity field. 

%% file: section/pilot.tex
\vspace{-1mm}
\paragraph{Subject-Driven Image Generation.} Subject-driven image generation aims to generate images conditioned on both a textual prompt, and a reference image, which provides the visual information defining the subject's unique identity. 
By incorporating visual reference information, subject-driven image generation not only enhances personalization and identity coherence but also allows the model to retain key appearance details of the subject, even when the scene undergoes significant changes.
Formally, we denote this image generation process as:
\begin{equation}
\mathbf{o} = \mathcal{G}(\mathbf{c}_{\text{prompt}}, \mathcal{I}_{\text{ref}}; \theta)
\end{equation}
where $\mathcal{G}$ is the generative model parameterized by \(\theta\), \(\mathbf{c}_{\text{prompt}}\) and \(\mathcal{I}_{\text{ref}}\) represent the textual prompt and the reference image, respectively. \(\mathbf{o}\) is the generated image.
In existing literature, \(\mathcal{G}\) is commonly implemented using diffusion models or flow matching models, both of which are based on iterative processes of noising and denoising.

\vspace{-1mm}
\paragraph{Optimization with Reinforcement Learning.}Recent work has framed this iterative visual generation process as a Markov Decision Process (MDP)~\citep{diffusion_rl} and employing reinforcement learning to maximize a given reward function~\citep{dancegrpo, flowgrpo, mixgrpo}.
Specifically, in the context of image generation, the MDP is a five-tuple $(S, A, \rho_0, P, R)$, each state $s_t \in S$ at timestep $t$ is represented as
$s_t \triangleq (c, t, z_t)$
where $c$ is the conditioning information (e.g., text prompt $\mathbf{c}_{\text{prompt}}$ and reference image $\mathcal{I}_{\text{ref}}$), $t$ is the current diffusion timestep, and $z_t$ denotes the corresponding noisy latent representation of the image.  
The action $a_t \in A$ corresponds to predicting the subsequent, less noisy latent $z_{t-\Delta t}$. 
The generative policy model $\mathcal{G}$ parameterized by $\theta$ serves as the the transition dynamics $P$ which determines the transition between latent states:
\begin{equation}
\pi_\theta(a_t \mid s_t) = \mathcal{G}_\theta(s_t),
\end{equation}
$\rho_0$ is the initial state distribution and $R$ is the reward function to measure the image quality.

Based on the MDP formulation, reinforcement learning, specifically Group Relative Policy Optimization (GRPO)~\citep{deepseek}, is employed to optimize the policy model $\pi_{\theta}$ by maximizing the following objective function:
\begin{equation} 
\resizebox{0.95\columnwidth}{!}{
  $\begin{displaystyle} 
  \begin{aligned} 
  \mathcal{J}(\theta) ={}& \mathbb{E}_{\substack{\{\mathbf{o}_i\}_{i=1}^G \sim \pi_{\theta_{\text{old}}}(\cdot|\mathbf{c}) \\ \mathbf{a}_{t,i} \sim \pi_{\theta_{\text{old}}}(\cdot|\mathbf{s}_{t,i})}} 
  \bigg[ \frac{1}{G} \sum_{i=1}^G \frac{1}{T} \sum_{t=1}^T \min\bigg( \rho_{t,i} A_i, \\ 
                 & \text{clip}\big( \rho_{t,i}, 1-\epsilon, 1+\epsilon \big) A_i \bigg) \bigg]
  \end{aligned}
  \end{displaystyle}$%
}
\label{eq:dancegrpoloss}
\end{equation}
where $\rho_t = \pi_\theta(a_t|s_t) / \pi_{\theta_{\text{old}}}(a_t|s_t)$ is the probability ratio between the new and old policies at timestep $t$ for sample $i$, and $\epsilon$ is the clipping hyperparameter to stabilize training. The expectation $\mathbb{E}$ is taken over groups of samples generated by the old policy $\pi_{\theta_{\text{old}}}$.
For a given condition $c$, generative models will sample a group of outputs $\{o_1, o_2, \dots, o_G\}$ from the model $\pi_{\theta_\text{old}}$.
Then, the advantage $A_i$ for each sample $o_i$ is then computed through intra-group normalization:
\begin{equation}
    A_i = \frac{r_i - \operatorname{mean}(\{ r_1, \ldots, r_G \})}{\operatorname{std}(\{ r_1, \ldots, r_G \}) + \epsilon_{\text{std}}}
\label{eq:advantage}
\end{equation}
where $r_i$ is the reward score for sample $o_i$, $\epsilon_{\text{std}}$ is a small constant to prevent division by zero. Following prior work~\citep{dancegrpo, flowgrpo}, we omit the KL-regularization term by default, as it yields minimal performance differences in our experiments.

\section{Pilot Experiments and Analysis}
\label{sec:pilot}

However, when applying reinforcement learning to subject-driven image generation, we encounter a key challenge: optimizing the policy model simultaneously with respect to two distinct objectives: \textbf{identity preservation} and \textbf{prompt adherence}. 
A straightforward approach is to aggregate these independent reward signals into a single composite reward function. 
In this formulation, the aggregated advantage for a batch of generated outputs $\{ o_1, o_2, \dots, o_G \}$ is expressed as: \begin{equation} A_{i}^{\text{naive}} = w_{\text{id}} \cdot A_{i}^{\text{id}} + w_{\text{prompt}} \cdot A_{i}^{\text{prompt}}, \label{eq:linear_advantages} \end{equation} where $A_{i}^{\text{id}}$ and $A_{i}^{\text{prompt}}$ denote the advantage derived from their respective reward $R_{\text{id}}$ and $R_{\text{prompt}}$. 
$w_{\text{id}}$ and $w_{\text{prompt}}$ are hyper-parameters that control the trade-off between the two objectives.


To evaluate the effectiveness of Eq.~\ref{eq:linear_advantages}, we conduct a series of preliminary experiments in this section.
Specifically, we adopt UNO~\citep{uno} as the base model and train it using Naive GRPO on a filtered subset of Syncd dataset~\citep{syncd}. We investigate two weighting strategies: a balanced strategy $(w_{\text{id}} : w_{\text{prompt}} = 1 : 1)$ and a prompt-biased configuration $(w_{\text{id}} : w_{\text{prompt}} = 1 : 1.5)$. We use DINO and HPS-v3~\citep{dinov2, hpsv3} as respective reward models for ID preservation and prompt following.

As shown in Figure~\ref{fig:hps_train}, when using the balanced weighting strategy, the model achieves a notable improvement in ID preservation, but at the cost of significantly impaired prompt-following capability. 
In contrast, the imbalanced weighting strategy ($w_{\text{id}} : w_{\text{prompt}} = 1 : 1.5$) alleviates the degradation in prompt-following objective, but substantially compromises the model's ability to preserve the subject identity.
Our experiments highlight a critical learning challenge: where optimizing for one objective (e.g., identity preservation) significantly conflicts with performance on the other (e.g., prompt following), leading to suboptimal overall subjective-driven image generation.

%% file: section/method.tex
\section{Method}
\label{sec:Method}

Building on the pilot analysis presented in Section~\ref{sec:pilot}, we propose \textbf{Customized-GRPO}, a variant of GRPO designed to overcome the limitations of linear advantage aggregation. Specifically, we introduce a \emph{synergy term} to mitigate reward conflicts (Section~\ref{sec:SARS}) and a \emph{dynamic weighting strategy} to address temporal misalignment (Section~\ref{sec:TDW}).

\subsection{Synergy-Aware Reward Shaping}
\label{sec:SARS}
As established in our pilot analysis, the linear aggregation of advantages is prone to suffer from signal cancellation under conflicting rewards. 
To address this, we introduce a non-linear term to shape the reward inspired by the principles of Pareto Optimization~\citep{multi_objective, self_improvement}. 
Instead of tolerating a trade-off, our goal is to guide the policy towards discovering solutions that the generated images excel in both ID preservation and prompt following simultaneously.

To address this issue, we introduce a \emph{synergy term} $\mathcal{S}$ that quantifies the alignment between the two advantage signals, $A_{\text{id}}$ and $A_{\text{prompt}}$. 
We define $\mathcal{S}$ via the hyperbolic tangent
\begin{equation}
    \mathcal{S} = \tanh\!\big(A_{\text{id}} \cdot A_{\text{prompt}}\big)
\end{equation}
which is bounded in $[-1,1]$ and saturates for large magnitudes, thereby acting as a stabilizing mechanism that limits gradient growth. 
The final advantage, $A^{\text{SARS}}$, is then computed as a piecewise aggregation based on the polarity of the individual advantages:
\begin{equation}  
\resizebox{\columnwidth}{!}{%
  $\displaystyle
  A^{\text{SARS}} = 
  \begin{cases} 
      w_{\text{id}} A_{\text{id}} + w_{\text{prompt}} A_{\text{prompt}} + \alpha \cdot \mathcal{S}, & \text{if } A_{\text{id}} > 0 \text{ or } A_{\text{prompt}} > 0, \\ 
      w_{\text{id}} A_{\text{id}} + w_{\text{prompt}} A_{\text{prompt}} - \alpha \cdot \mathcal{S}, & \text{if both } A_{\text{id}} \leq 0 \text{ and } A_{\text{prompt}} \leq 0, 
  \end{cases}
  $
}
\label{eq:SARS}
\end{equation}
where $\alpha \ge 0$ controls the influence of the synergy term. 
This formulation yields a more informative learning signal than simple linear aggregation.

\begin{figure}[htbp]
  \centering
  \setlength{\abovecaptionskip}{0.2cm}
  \setlength{\belowcaptionskip}{-0.4cm}
  \includegraphics[width=0.45\textwidth]{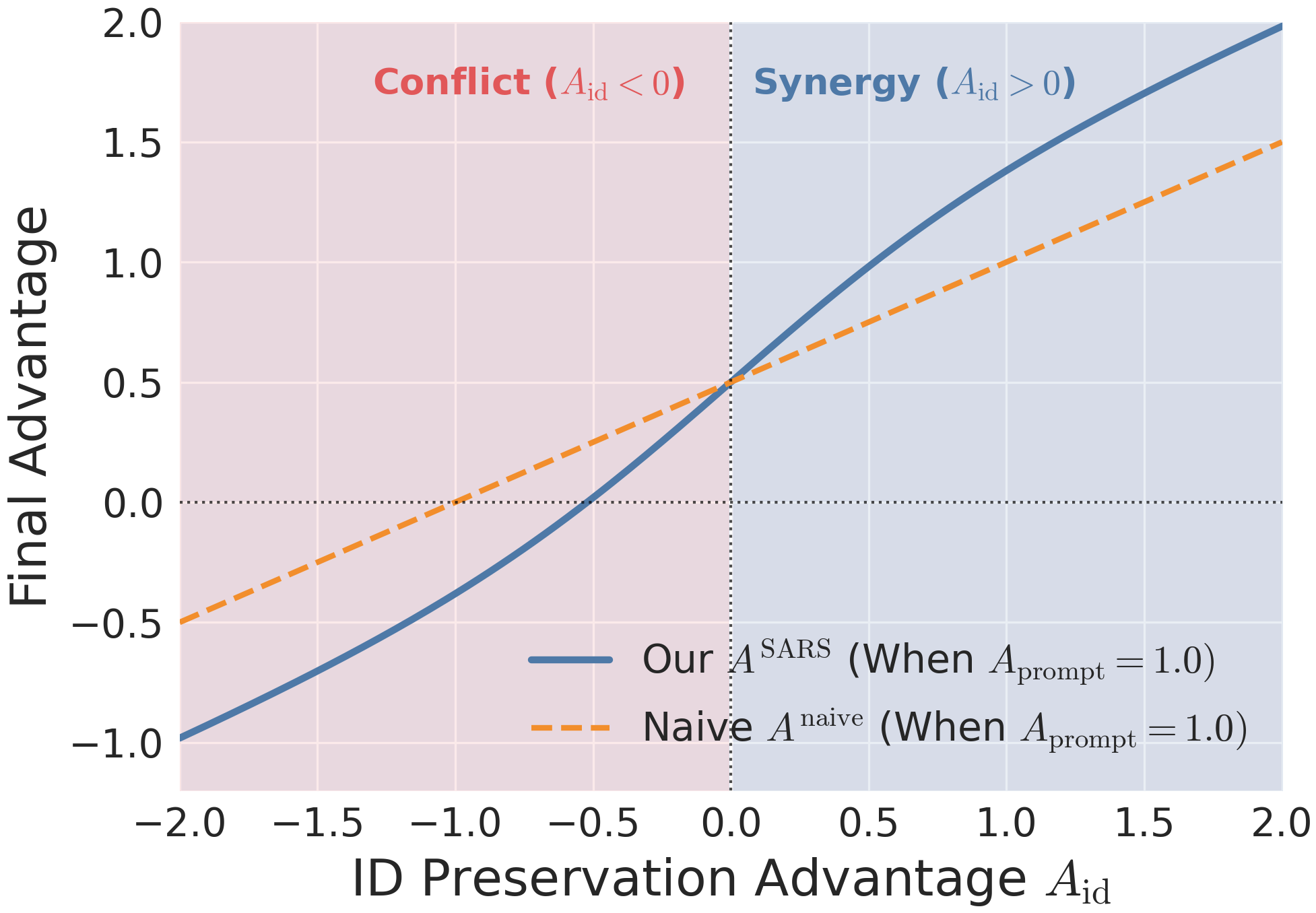}
  \caption{\textbf{Cross-section of the SARS function, illustrating how it reshapes the linear reward landscape.} We plot the final advantage as a function of $A_{\text{id}}$ while holding $A_{\text{prompt}}$ constant at 1.0.}
  \label{fig:sars_cross}
\end{figure}


As visualized in Figure~\ref{fig:sars_cross}, our SARS function (solid curve) fundamentally reshapes the naive linear baseline (dashed curve). It non-linearly amplifies the reward in the Synergy region ($A_{\text{id}}>0$) and applies a decisive penalty in the Conflict region ($A_{\text{id}}<0$). This non-linear transformation creates a sharper gradient around the decision boundary.

By incorporating the synergy term, our method replaces ambiguous, linearly combined feedback with sharper, decision-consistent signals. 
This alleviates cancellation and better aligns optimization with the goal of producing well-balanced, high-quality images. 

\begin{figure}[htbp]
  \centering
  \setlength{\abovecaptionskip}{0.2cm}
  \setlength{\belowcaptionskip}{-0.4cm}
  \includegraphics[width=0.5\textwidth]{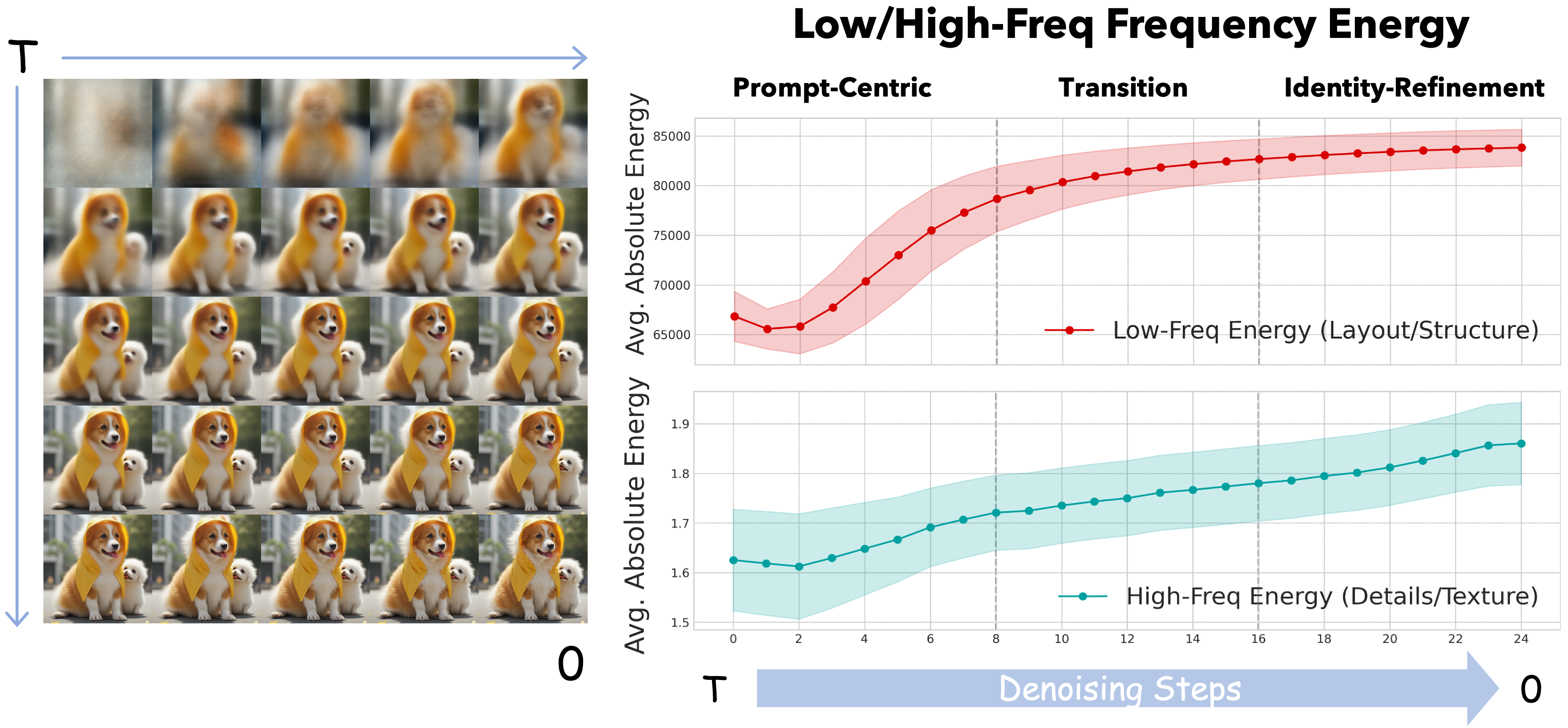}
  \caption{\textbf{Temporal Dynamics of Feature Synthesis during the Denoising Process}. (Right) The plots show that low-frequency energy (structure) converges early, whereas high-frequency energy (details) accumulates steadily over all phases. (Left) The visual progression from pure noise (t=T) to the final image (t=0) corroborates this finding.}
  \label{fig:timesteps}
  \vspace{-3mm}
\end{figure}

\subsection{Timestep-Aware Dynamic Weighting}
\label{sec:TDW}

Previous work has established that the diffusion denoising process is not static but exhibits distinct functional phases~\citep{denoising, blip_diffusion, progressive, ediff}. As confirmed by our FFT analysis in Figure~\ref{fig:timesteps}, the early stages are dominated by the formation of low-frequency global structure, presenting an opportune window to prioritize prompt following as the model makes high-level compositional decisions. Conversely, the later stages are characterized by the steady accumulation of high-frequency local details, making this phase critical for refining ID preservation. Static weighting strategy, however, suffers from Temporal Misalignment, fundamentally conflicting with the dynamic nature of the denoising process.

To address this, we propose Timestep-Aware Dynamic Weighting (TDW), which replaces static optimization with a curriculum-inspired strategy that adapts weighting over time. The core idea is to align the optimization pressure with the model's transient focus. We partition the denoising trajectory into three distinct stages: an early Prompt-Centric Phase to establish global structure, a late Identity-Refinement Phase to perfect local details, and a smooth Transition Phase bridging them.



Specifically, we design the following function for $w_{\text{prompt}}(t)$:
\begin{equation}
\resizebox{\columnwidth}{!}{%
  $\begin{displaystyle}
  w_{\text{prompt}}(t) =
  \begin{dcases*}
      w_{\text{max}} & if $T \ge t > T_{\text{trans}}$ \\
      \begin{aligned}[t]
          & w_{\text{min}} + (w_{\text{max}} - w_{\text{min}}) \cdot \\
          & \qquad (1 - \sigma(\lambda(t - t_c)))
      \end{aligned} & if $T_{\text{trans}} \ge t > T_{\text{id}}$ \\
      w_{\text{min}} & if $T_{\text{id}} \ge t > 0$
  \end{dcases*}
  \end{displaystyle}$%
}
\label{eq:dynamic_weights}
\end{equation}
where the weight for ID preservation is $w_{\text{id}}(t) = 1 - w_{\text{prompt}}(t)$. To ensure stability, the weights are constrained by upper and lower bounds $w_{\text{max}}$ and $w_{\text{min}}$ and a Sigmoid function $\sigma(\cdot)$ provides a smooth interpolation during the transition phase, preventing abrupt policy shifts.

By temporally decoupling the competing objectives, TDW enables a more principled optimization. 
It transforms the learning problem from navigating a static, inefficient trade-off into a curriculum-like process, allowing the policy to allocate its limited gradient budget to the most critical task at each timestep.

%% file: section/experiments.tex
\section{Experiments}
\subsection{Experiment Setup}
\paragraph{Implementation details.} Our main experiments are conducted on the subject-driven generation model UNO~\citep{uno}, a Diffusion Transformer (DiT) based architecture built upon FLUX.1-dev~\citep{flux}. We freeze the pretrained DiT backbone and the image encoder, training only the injected LoRA~\citep{lora} matrices. We provide training details in Appendix~\ref{app:experimental}.

\paragraph{Dataset and Reward Models.} We use a high-quality, filtered subset of 10k samples from the large-scale Syncd dataset~\citep{syncd} for training. To guide our policy optimization, we define two reward functions corresponding to our core objectives. For ID preservation, 
we use a segmentation-masked DINOv2~\citep{dinov2} score, termed DINO-Seg, to measure the fidelity between the subject in the generated image and the reference. It remove the background variations when computing the image similarity to better reflect the faithfulness to the reference subject. For prompt following, we use the Human Preference Score v3 (HPS-v3)~\citep{hpsv3}, a powerful reward model trained on a large dataset of human preferences to evaluate alignment with the text prompt.

\paragraph{Evaluation Protocol.} We conduct a comprehensive evaluation on the widely-used DreamBench~\citep{dreambooth}. Our method is compared against a suite of state-of-the-art baselines, including both fine-tuning-based methods: Textual Inversion~\citep{textual_inversion}, DreamBooth~\citep{dreambooth}; finetuning-free methods BLIP-Diffusion~\citep{blip_diffusion}, ELITE~\citep{elite}, SSR-Encoder~\citep{ssr_encoder}, OminiControl~\citep{ominicontrol}, OmniGen~\citep{omnigen}, FLUX IP-Adatper~\citep{flux_ipadapter} and RL methods RPO~\cite{subject_rl}. Performance is measured using established metrics: CLIP-T~\citep{clipt} for prompt adherence, CLIP-I and DINO Score~\citep{clipscore,dino} for identity preservation, and the HPS-v3~\citep{hpsv3} for text-image alignment and aesthetics.




\subsection{Main Results}
Table~\ref{tab:quantitative} reports the main quantitative results of our method and against state-of-the-art baselines on Dreambench. 
Our Customized-GRPO demonstrates superior overall performance. It achieves the highest scores with a DINO score of 0.812 and a CLIP-I score of 0.872. While the CLIP-T score (0.301), which measures strict text-image similarity, remains on par with top-performing baselines like OmniGen (0.315) and DreamBooth (0.305), our method attains the highest Human Preference Score (HPS-v3) of 8.32, which evaluates both text-image alignment and overall aesthetic quality. It indicates that while maintaining a strong textual alignment, our policy optimization has successfully learned to generate images with higher overall aesthetic quality and are better aligned with human perception.

Overall, Customized-GRPO improves upon our UNO base model by 6.8\% in the DINO score for ID preservation while simultaneously increasing the HPS-v3 score for prompt following and human preference by 5.7\%. The concurrent improvement across both ID preservation and prompt following metrics provides clear evidence that our method effectively mitigates the competitive degradation observed in naive approaches and successfully learns a more balanced policy.

We conduct a pairwise human evaluation comparing Customized-GRPO against the base UNO model. As shown in Figure~\ref{fig:user_study}, our method is strongly preferred by human annotators, winning in Prompt Following and ID Preservation by decisive margins of 50\% vs. 22\% and 54\% vs. 21\%, respectively. In both criteria, our method is preferred more than twice as often as the base model.

We provide qualitative comparisons in Appendix~\ref{app:qualitative}, which further visualize our method's superior ability to preserve identity while adhering to complex prompts.

\begin{table}[t]
    \centering
    \setlength{\abovecaptionskip}{0.2cm}
    \setlength{\belowcaptionskip}{-0.4cm}
    \resizebox{0.48\textwidth}{!}{
        \small
        \begin{tabular}{l|cccc}
            \toprule
            \textbf{Method} & \textbf{DINO$\uparrow$} & \textbf{CLIP-I$\uparrow$} & \textbf{CLIP-T$\uparrow$} & \textbf{HPS-v3$\uparrow$}\\
            \midrule
            Textual Inversion         & 0.569    & 0.780    & 0.255 & -- \\
            DreamBooth                & 0.668    & 0.803    & 0.305 & -- \\
            BLIP-Diffusion            & 0.670    & 0.805    & 0.302 & -- \\
            ELITE                     & 0.647    & 0.772    & 0.296 & 3.12 \\
            SSR-Encoder               & 0.612    & 0.821    & 0.308 & 4.82 \\
            OmniGen                   & 0.693    & 0.801    & \textbf{0.315} & 6.67 \\
            OminiControl               & 0.684    & 0.799    & 0.312 & 8.28 \\
            FLUX.1 IP-Adapter         & 0.582    & 0.820    & 0.288 & 7.68 \\
            \midrule
            UNO (Base)       & 0.760 & 0.835 & 0.304 & 7.87 \\
            UNO (RPO) & 0.852 & 0.898 & 0.237 & 4.47 \\
            \textbf{UNO (Customized-GRPO)} & \textbf{0.812} & \textbf{0.862} & 0.301 & \textbf{8.32} \\
            \bottomrule
        \end{tabular}
    }
    \caption{Quantitative Results for subject-driven generation on Dreambench.}
    \label{tab:quantitative}
\end{table}

\begin{table}[t]
    \centering
    \setlength{\abovecaptionskip}{0.2cm}
    \setlength{\belowcaptionskip}{-0.4cm}
    \resizebox{0.48\textwidth}{!}{
        \small
        \begin{tabular}{l|cccc}
            \toprule
            \textbf{Method} & \textbf{DINO$\uparrow$} & \textbf{CLIP-I$\uparrow$} & \textbf{CLIP-T$\uparrow$} & \textbf{HPS-v3$\uparrow$}\\
            \midrule
            UNO (Base) & 0.760 & 0.835 & 0.304 & 7.87 \\
            \midrule
            SFT & 0.762 & 0.846 & \textbf{0.307} & 7.43 \\
            RPO & 0.852 & 0.898 & 0.237 & 4.47 \\
            \rowcolor{red!40}
            Naive GRPO (1:1) & 0.861 & 0.912 & 0.235 & 4.12 \\
            Naive GRPO (1:1.5) & 0.801 & 0.842 & 0.298 & 7.56 \\
            \midrule
            Customized-GRPO & \textbf{0.812} & \textbf{0.862} & 0.301 & \textbf{8.32} \\
            w/o Synergy Term & 0.795 & 0.838 & 0.295 & 7.95 \\
            w/o TDW & 0.811 & 0.852 & 0.278 & 7.23 \\ 
            \bottomrule
        \end{tabular}
    }
    \caption{Ablation Results on Dreambench. The best performance is marked in \textbf{bold}. We highlight the row in \textcolor{red!60}{red} to denote the competitive degradation.}
    \label{tab:ablation}
\end{table}

\subsection{Ablation Study}
We conduct ablation studies to validate the efficacy of our method: Synergy-Aware Reward Shaping and Time-Aware Dynamic Weighting. We compare our full Customized-GRPO method against ablated versions where each component is removed. The results are presented in Table~\ref{tab:ablation}.
\paragraph{Effect of Synergy-Aware Reward Shaping.}
We evaluate a model with only TDW active (w/o Synergy Term), which applies dynamic weights but relies on a simple linear advantage sum. It successfully prevents the catastrophic performance collapse seen in prompt following observed in the Naive GRPO ( HPS-v3 = 7.95 vs. 4.12) but all metrics remain notably lower than our full model. This indicates that while TDW correctly allocates optimization pressure across timesteps, it is insufficient to resolve the underlying reward conflict at each individual output in the generated group. The synergy term is therefore critical for providing a sharper, more decisive learning signal that leads to a superior final policy.
\paragraph{Effect of Time-Aware Dynamic Weighting.}
We evaluate a model with only SARS active (w/o TDW), which uses our conflict-aware advantage function with fixed 1:1 weights across all timesteps. While this model also improves upon the naive baseline by penalizing conflicted outputs, its performance on prompt following metrics is substantially weaker than full method. This result confirms our hypothesis: without dynamically shifting the focus from prompt following to ID preservation, the static optimization pressure still leads to an imbalanced policy that over-emphasizes fidelity in the later, detail-focused stages of denoising.
\begin{figure}[t]
    \centering
    \setlength{\abovecaptionskip}{0.2cm}
    \setlength{\belowcaptionskip}{-0.4cm}
    \includegraphics[width=0.45\textwidth]{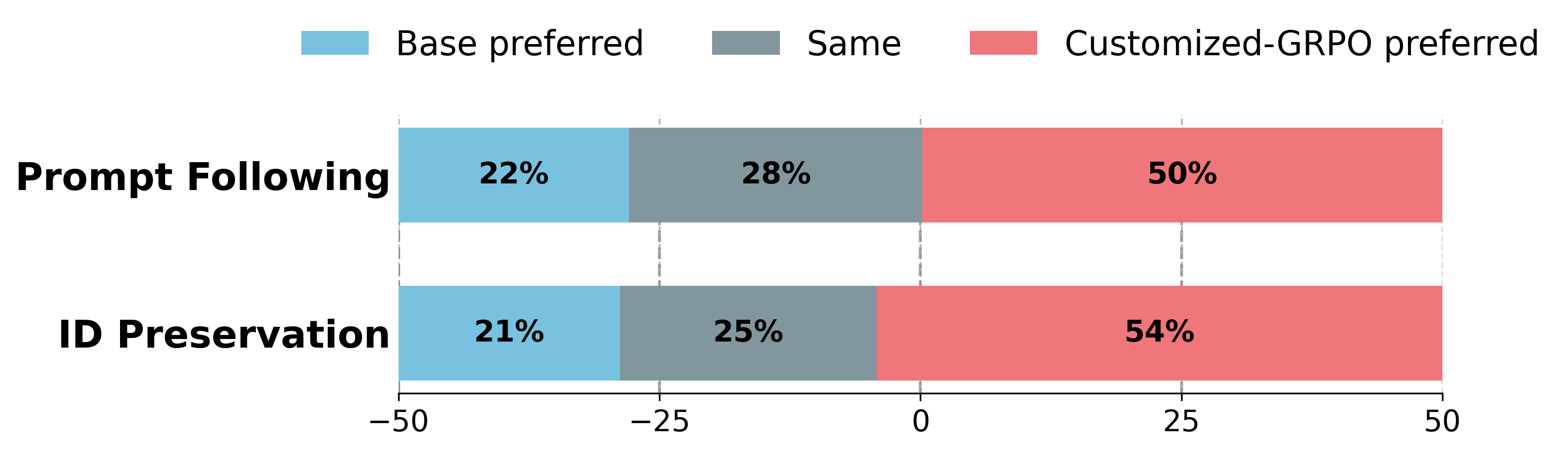}
    \caption{\textbf{Human evaluation.} Pairwise comparison between our Customized-GRPO and the base model.}
    \label{fig:user_study}
\end{figure}

\begin{table}[t]
    \centering
    \setlength{\abovecaptionskip}{0.2cm}
    \setlength{\belowcaptionskip}{-0.4cm}
    \resizebox{0.48\textwidth}{!}{
        \small
        \begin{tabular}{l|cccc}
            \toprule
            \textbf{Method} & \textbf{DINO$\uparrow$} & \textbf{CLIP-I$\uparrow$} & \textbf{CLIP-T$\uparrow$} & \textbf{HPS-v3$\uparrow$}\\
            \midrule
            UNO (Base) & 0.760 & 0.835 & \textbf{0.304} & 7.87 \\
            \midrule
            Naive GRPO & 0.801 & 0.842 & 0.298 & 7.56 \\
            Min($A_{\text{id}},A_{\text{prompt}}$) & 0.785 & 0.848 & 0.298 & 7.81 \\
            \rowcolor{red!40}
            Max($A_{\text{id}},A_{\text{prompt}}$) & 0.884 & 0.899 & 0.276 & 6.26 \\
            Harmonic Mean & 0.778 & 0.812 & 0.291 & 8.22\\ 
            \midrule
            \textbf{Tanh($A_{\text{id}} \cdot A_{\text{prompt}}$)} & \textbf{0.812} & \textbf{0.862} & 0.301 & \textbf{8.32} \\
            \bottomrule
        \end{tabular}
    }
    \caption{Analysis on the choice of Synergy Term. The best performance is marked in \textbf{bold}. We highlight the row in \textcolor{red!60}{red} to denote the competitive degradation.}
    \label{tab:synergy_term}
\end{table}

\begin{table}[t]
    \centering
    \setlength{\abovecaptionskip}{0.2cm}
    \setlength{\belowcaptionskip}{-0.4cm}
    \resizebox{0.48\textwidth}{!}{
        \small
        \begin{tabular}{lcccc}
            \toprule
            \textbf{$w_{max}:w_{min}$} & \textbf{DINO$\uparrow$} & \textbf{CLIP-I$\uparrow$} & \textbf{CLIP-T$\uparrow$} & \textbf{HPS-v3$\uparrow$}\\
            \midrule
            $1.0:0$ & 0.832 & 0.874 & 0.272 & 6.39 \\
            $0.9:0.1$ & 0.821 & 0.866 & 0.285 & 7.27 \\
            $0.8:0.2$ & 0.814 & 0.864 & 0.296 & 7.68 \\
            $\boldsymbol{0.7:0.3}$ & \textbf{0.812} & \textbf{0.862} & \textbf{0.301} & \textbf{8.32} \\
            $0.6:0.4$ & 0.795 & 0.848 & 0.292 & 7.88 \\
            \bottomrule
        \end{tabular}
    }
    \caption{Analysis on the choice of Dynamic Weight. The best performance is marked in \textbf{bold}.}
    \label{tab:dynamic_weight}
\end{table}

\begin{figure}[t]
    \centering
    \setlength{\abovecaptionskip}{0.2cm}
    \setlength{\belowcaptionskip}{-0.4cm}
    \includegraphics[width=0.45\textwidth]{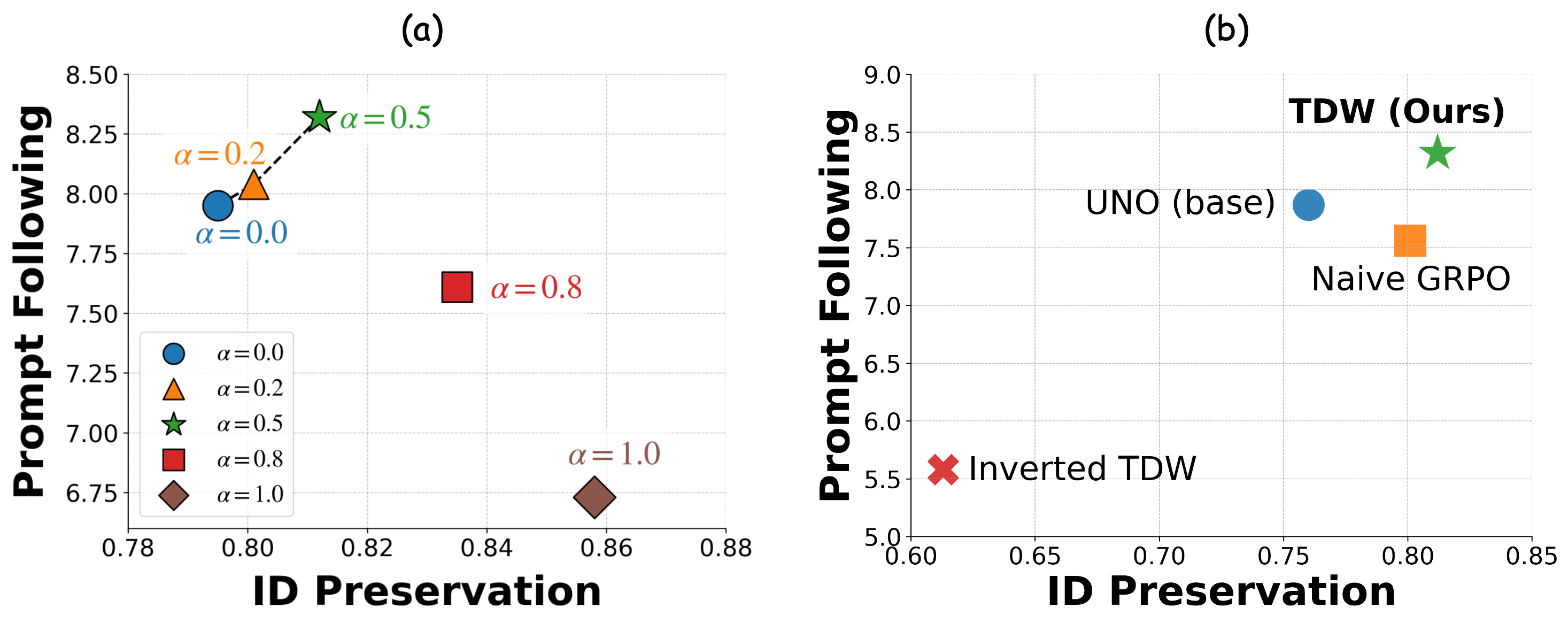}
    \caption{Analysis of the Synergy Coefficient and TDW Schedule.}
    \label{fig:analyses}
    \vspace{-2mm}
\end{figure}
\subsection{Analyses}
\label{sec:analyses}
\paragraph{Analysis on the choice of Synergy Term.}
We conduct a comparative analysis of linear function ($w_{\text{id}}:w_{\text{prompt}}=1:1.5$) and four non-linear functions to determine the most effective function for synergy term. The results are presented in Table~\ref{tab:synergy_term}.
Max function exhibits classic reward hacking, achieving high fidelity scores by sacrificing prompt following, where the model learns to maximize one objective at the severe expense of the other. Min function, conversely, proves too conservative and fails to deliver significant improvements.

While Harmonic Mean and $\text{Tanh}$ functions both designed to encourage balance, yielding promising results by successfully avoiding reward hacking, $\text{Tanh}$ function proves superior, attaining the highest HPS-v3 score of 8.32 while maintaining a strong, balanced performance across all metrics.

We attribute the success to its ideal mathematical properties for this task: its bounded and symmetric nature provides a stable and consistent signal. Therefore, we adopt it as the synergy function.
\paragraph{Comparison with RL and SFT.}
To further validate the efficacy of our online RL approach, we conduct a direct comparison against Supervised Fine-Tuning (SFT) and a recent offline preference optimization baseline, RPO~\cite{subject_rl}. We fine-tune the UNO (base) model using SFT and RPO on the same 10k high-quality data curated for our experiments. The results, presented in Table~\ref{tab:ablation}, show that SFT yields only marginal or even negative changes compared to the base model.  Since the base model is already well-trained on a similar data distribution, simply continuing to imitate this static dataset offers no clear path to resolving the complex trade-off. Conversely, while offline RPO significantly improves identity fidelity (DINO: 0.852), it incurs severe competitive degradation in text alignment, with the HPS-v3 score dropping drastically to 4.47. This indicates that RPO tends to overfit the reference image, failing to balance conflicting objectives due to its static harmonic mean aggregation and the inherent challenge of constructing balanced offline preference pairs (implementation details in Appendix~\ref{app:experimental}).
In contrast, our Customized-GRPO method demonstrates significant performance gains across both objectives.

\paragraph{Dynamic Weight Bounds Analysis.}
For dynamic weight bounds $w_{max}, w_{min}$, we evaluate the impact of different dynamic ranges on model performance while keeping other hyper-parameters fixed. The results are presented in Table~\ref{tab:dynamic_weight}. Our observations reveal two notable trends: \textbf{overly aggressive decay} ($w_{min}\le0.1$) causes \textit{semantic collapse} in the final denoising stages, where the model tends to overfit the reference image and fails to adhere to textual constraints. Conversely, \textbf{insufficient dynamic range}($w_{max}:w_{mian} = 0.6:0.4$) fails to resolve competitive degradation, resulting in drops for both DINO (0.795) and HPS-v3 (7.88) compared to the optimal setting. Based on these results, we adopt $w_{max}:w_{mian} = 0.7:0.3$ as the default configuration in this work.

\paragraph{Synergy Coefficient Analysis.}
We analyze the impact of the synergy coefficient $\alpha$, which controls the strength of the SARS adjustment. As shown in Figure~\ref{fig:analyses} (a), the results reveal a clear optimal balance. When $\alpha$ increase from 0 to 0.5, we observe a steady improvement on both objectives, demonstrating the effectiveness of SARS in transforming conflicted signals into a productive gradient.
However, further increasing $\alpha$ to 0.8 and 1.0 leads to a significant performance degradation. This indicates that an overly large $\alpha$ induces a risk-averse policy that excessively penalizes conflict, thereby failing to achieve a good balance. Based on this analysis , we adopt $\alpha=0.5$ for all subsequent experiments.


\paragraph{Validation of TDW Weighting Schedule.}
We conduct an analysis comparing our standard TDW against two critical ablations: a Naive GRPO using static weights, and an Inverted TDW that reverses the curriculum. As shown in Figure~\ref{fig:analyses}(b), the Inverted TDW suffers a severe collapse in performance, empirically confirming that our schedule is essential and aligns with the model's coarse-to-fine generation process. 
Meanwhile, Naive GRPO demonstrates the inefficient trade-off inherent in static weighting. Our standard TDW is positioned distinctly in the top-right corner, demonstrating its superior ability to mitigate the trade-off and successfully improve both objectives simultaneously.

\begin{table}[t]
    \centering
    \setlength{\abovecaptionskip}{0.2cm}
    \setlength{\belowcaptionskip}{-0.4cm}
    \resizebox{0.48\textwidth}{!}{
        \begin{tabular}{c|cc|cc}
            \toprule
            \multirow{2}{*}{\textbf{Synergy Coef. ($\alpha$)}} & \multicolumn{2}{c|}{\textbf{Standard TDW}} & \multicolumn{2}{c}{\textbf{Inverted TDW}} \\
            \cmidrule{2-5}
            & \textbf{DINO$\uparrow$} & \textbf{HPS-v3$\uparrow$} & \textbf{DINO$\uparrow$} & \textbf{HPS-v3$\uparrow$} \\
            \midrule
            $\alpha = 0.0$ & 0.795 & 7.95 & 0.605 & 5.65 \\
            $\alpha = 0.2$ & 0.801 & 8.04 & 0.609 & 5.62 \\
            $\boldsymbol{\alpha = 0.5}$ & \textbf{0.812} & \textbf{8.32} & \textbf{0.613} & \textbf{5.58} \\
            $\alpha = 0.8$ & 0.835 & 7.61 & 0.616 & 5.49 \\
            $\alpha = 1.0$ & 0.858 & 6.73 & 0.620 & 5.41 \\
            \bottomrule
        \end{tabular}
    }
    \caption{Analysis of the interaction between Synergy Coefficient and TDW schedules.}
    \label{tab:interaction_analysis}
\end{table}

\paragraph{Interaction between SARS and TDW.}
To systematically analyse the hyperparameter interactions within our framework, we find that SARS and TDW address competitive degradation from distinct dimensions. We propose that TDW acts as the performance foundation, while SARS acts as the performance amplifier. Consequently, the efficacy of SARS depends on a reasonable TDW schedule. To validate this, we conduct comprehensive experiments on DreamBench, varying $\alpha \in 0.0,0.2,0.5,0.8,1.0$ across both the standard TDW schedule and the Inverted TDW schedule (which reverses the curriculum). 

As shown in Table~\ref{tab:interaction_analysis}, aligning the weighting schedule with diffusion dynamics is the foundation for success. An inverted schedule leads to a performance collapse (DINO $0.6$, HPS-v3 $0.3$) regardless of the $\alpha$ value, showing minimal sensitivity to parameter changes.
Under the correct Standard TDW, when $\alpha$ increases from 0 to 0.5, we observe a steady improvement on both objectives. However, further increasing $\alpha$ to 0.8 and 1.0 triggers \textbf{reward hacking}, where the model over-optimizes Identity at the significant expense of Prompt adherence.
\section{Conclusion}
In this work, we introduce Customized-GRPO, a novel reinforcement learning framework featuring Synergy-Aware Reward Shaping (SARS) and Time-Aware Dynamic Weighting (TDW) to resolve the critical fidelity-editability trade-off in subject-driven generation. Experiments show that naive GRPO approaches fail due to reward conflict and temporal misalignment, while our method successfully generates images that are simultaneously faithful to the subject's identity and accurately adhere to complex prompts.

\section*{Limitation}
In this section, we discuss the limitation of our work:
(1) Due to computational constraints, our experiments are conducted exclusively on a Diffusion Transformer (DiT) architecture. While our framework is theoretically model-agnostic, future work should involve applying Customized-GRPO to other generative architectures, particularly auto-regressive models, to fully validate its generalization across different modeling paradigms.
(2) Our current framework focuses on optimizing the generation of a single subject per image. However, a significant emerging challenge in personalized generation is multi-concept composition, which involves generating an image containing multiple, distinct subjects. We will extend our reinforcement learning framwork to manage and balance the rewards for multiple, interacting concepts in the future work.

\section*{Ethics Statement}
The training data used in our experiments is a filtered subset of the publicly available dataset. During the curation process, we applied automated filtering mechanisms to detect and remove any images containing material, or personally identifiable information. For the textual prompts, we conduct a comprehensive manual review to further ensure the appropriateness of the content. 

\section*{Acknowledgment}
This work was supported in part by the Ningbo Youth Science and Technology Innovation Leading Talent Program (No. 2025QL059), and the Earth System Big Data Platform of the School of Earth Sciences, Zhejiang University.

%% file: section/appendix.tex
\section{Details on Experimental Setup}
\label{app:experimental}

\subsection{Implementation Details}
Our main experiments are conducted on the subject-driven generation model UNO~\citep{uno}, a Diffusion Transformer (DiT) based architecture built upon FLUX.1-dev~\citep{flux}. We freeze the pretrained DiT backbone and the image encoder, training only the injected LoRA~\citep{lora} matrices.
For training-time sampling, we set T = 25 as the total sampling steps. For GRPO, the model enerates 12 images for each prompt and clips the advantage to the range [1e-5, 1e-5]. We use AdamW  as the optimizer with a learning rate of 1e-5 and a weight decay coefficient of 0.0001. 
All experiments are conducted on 8 NVIDIA A100 GPUs with a batch size of 1, for a maximum of 1250 training steps, resulting in a total of approximately 80 wall-clock hours (i.e., 640 GPU hours).
\paragraph{Hyperparameter Configuration.}
For Synergy-Aware Reward Shaping (SARS), we set the synergy coefficient $\alpha$ to 0.5 based on the ablation study in Section~\ref{sec:analyses}. For Time-Aware Dynamic Weighting (TDW), the weights for prompt following, $w_{\text{prompt}}(t)$, are bounded by $w_{\text{max}} = 0.7$ and $w_{\text{min}} = 0.3$. The three optimization phases are partitioned as follows: Prompt-Centric Phase for $t < 6$, Transition Phase for $6 \le t < 22$, and Identity-Refinement Phase for $t \ge 22$.

\paragraph{Training Dataset.} We utilize the Syncd dataset~\citep{syncd}, a large-scale dataset designed for subject-driven generation, including rigid and deformable categories. To create a more focused and high-quality subset for our experiments, we filter 10,000 datasets with the highest average DINOv2 similarity and aesthetic scores.

\paragraph{Reward Model.} For ID Preservation, we use DINOv2 (ViT-L/14)~\citep{dinov2} as the foundation for our fidelity reward. To ensure that the reward signal is focused specifically on the subject and not influenced by background, we employ a object detection and segmentation approach. This helps remove the  background variations when computing the image similarity scores to better reflect the faithfulness to the reference  subject. We call it DINO-Seg. For Prompt Following, we use the Human Preference Score v3 (HPSv3)~\citep{hpsv3} as our reward model. HPSv3 is a powerful reward model trained on a large dataset of human preferences, and it provides a reliable score that reflects how well a generated image aligns with the semantics of a given text prompt.

\subsection{Evaluation Details}
\paragraph{Benchmark.} To ensure a robust and comprehensive assessment, we evaluate all methods on the widely-used DreamBench~\citep{dreambooth}. For each of the 30 subjects in the benchmark, we use all associated prompts. To ensure evaluation stability and account for stochasticity in the generation process, we generate 4 images per prompt and report the average score across all generated images.

\paragraph{Baselines.} We compare our method with SOTA methods including both finetuning-based methods: Textual Inversion~\citep{textual_inversion}, DreamBooth~\citep{dreambooth} and finetuning-free methods: BLIP-Diffusion~\citep{blip_diffusion}, ELITE~\citep{elite}, SSR-Encoder~\citep{ssr_encoder}, OminiControl~\citep{ominicontrol}, OmniGen~\citep{omnigen} and FLUX IP-Adatper~\citep{flux_ipadapter}.

Regarding the comparison with RPO~\cite{subject_rl}, given that their original data is not publicly available, we strictly replicate their data construction pipeline to curate a paired preference dataset using our own training data~\cite{syncd}. Specifically, we employ the frozen UNO base model to generate candidate images conditioned solely on text prompts. Subsequently, we adopt RPO’s core Harmonic Mean reward aggregation mechanism to distinguish between winning and losing samples for preference pair construction. Following RPO protocol, we fine-tune UNO using the composite objective proposed, which combines the Offline DPO~\cite{dpo} loss for preference optimization with a Similarity Loss to ensure identity consistency.

We adopted the default hyperparameters specified for each model by their respective authors.

\paragraph{Evaluation Metrics.} We follow previous methods to adopt three metrics (CLIP-T, CLIP-I, and DINO) for evaluation. Specifically, CLIP-T evaluates the similarity between the generated images and given text prompts; CLIP-I and DINO evaluate the similarity between the generated images and the reference images. To better capture overall quality and alignment with human perception, we additionally report the HPS-v3~\citep{hpsv3}. This metric evaluates not only text-image alignment but also broader aesthetic qualities. 

\subsection{Human Evaluation Details}
To complement our quantitative metrics, we conduct a comprehensive human evaluation comparing our Customized-GRPO against base UNO model. We randomly select 200 diverse subject-prompt pairs from DreamBench, generating one image per model (Customized-GRPO and UNO base) for each pair. We engage eight experienced, English-fluent annotators on iTAG\footnote{https://www.alibabacloud.com/help/en/pai/user-guide/itag/} platform for the task. Before evaluation, all annotators completed a detailed tutorial to align their understanding of the criteria. As illustrated in Figure~\ref{fig:user_interface}, the evaluation interface present annotators with the text prompt and two sets of images. For each model being compared (Model 1 and Model 2), we display its generated image alongside the corresponding reference image. The positions of Model 1 and Model 2 were randomized to prevent positional bias. Annotators perform a pairwise comparison, answering two question: "which model performs better based on the visual similarity between the generated image and the reference image of the object/animal?" and  "which model performs better by evaluating both (1) how well the generated image align with the provided prompt, and (2) the overall visual quality (aesthetics) of the generated image?" For each question, annotators could choose one of the two models or select "Tie" if they are of similar quality. 

\begin{algorithm*}[htbp]
\caption{Customized-GRPO Training for Subject-Driven Generation}\label{algo:customized_grpo}
\small
\begin{algorithmic}[1]
\Require Initial policy model $\pi_\theta$; ID reward model $R_{\text{id}}$; Prompt reward model $R_{\text{prompt}}$;
\Require Paired dataset $\mathcal{D} = \{(\mathcal{I}_j, \mathbf{c}_j)\}_{j=1}^N$ of reference images and prompts;
\Require Total sampling steps $T$; Synergy hyperparameter $\alpha$;
\Require Dynamic weighting function $f(t) \to (w_{\text{id}}(t), w_{\text{prompt}}(t))$

\Ensure Optimized policy model $\pi_\theta$

\For{training iteration $=1$ \textbf{to} $M$}
    \State Sample a batch of (reference image, prompt) pairs $\{(\mathcal{I}_j, \mathbf{c}_j)\}_{j=1}^B \sim \mathcal{D}$
    \State Update old policy: $\pi_{\theta_{\text{old}}} \gets \pi_\theta$
    
    \For{each pair $(\mathcal{I}, \mathbf{c})$ in the batch}
        \State Generate a group of $G$ outputs: $\{\mathbf{o}_i\}_{i=1}^G \sim \pi_{\theta_{\text{old}}}(\cdot|\mathcal{I}, \mathbf{c})$
        \State Compute advantages $A^{\text{id}}, A^{\text{prompt}}$ for each sample $i$: $A_i \gets \sum_{k=1}^K \frac{r_i^k - \mu^k}{\sigma^k}$
        
        \For{each output $i \in \{1, \dots, G\}$}
            \State $\mathcal{S}_i \gets \tanh(A_i^{\text{id}} \cdot A_i^{\text{prompt}})$ \Comment{Synergy-Aware Shaping}
        \EndFor

        \State Subsample a subset of training timesteps $\mathcal{T}_{\text{sub}} \subset \{1, \dots, T\}$
        
        \For{$t \in \mathcal{T}_{\text{sub}}$}
            \State $(w_{\text{id}}(t), w_{\text{prompt}}(t)) \gets f(t)$ \Comment{Timestep-Aware Weighting}
            
            \For{each output $i \in \{1, \dots, G\}$}
                \If{$A_i^{\text{id}} > 0$ \textbf{or} $A_i^{\text{prompt}} > 0$}
                    \State $A_i^{\text{final}}(t) \gets w_{\text{id}}(t)A_i^{\text{id}} + w_{\text{prompt}}(t)A_i^{\text{prompt}} + \alpha \cdot \mathcal{S}_i$
                \Else
                    \State $A_i^{\text{final}}(t) \gets w_{\text{id}}(t)A_i^{\text{id}} + w_{\text{prompt}}(t)A_i^{\text{prompt}} - \alpha \cdot \mathcal{S}_i$
                \EndIf \Comment{Combine synergy and dynamic weights}
            \EndFor
            
            \State Compute GRPO objective $\mathcal{J}$ using time-dependent advantages $\{A_i^{\text{final}}(t)\}_{i=1}^G$
            \State Update policy parameters via gradient ascent: $\theta \gets \theta + \eta \nabla_\theta \mathcal{J}$
        \EndFor
    \EndFor
\EndFor
\State \Return $\pi_\theta$
\label{algo:customized}
\end{algorithmic}
\end{algorithm*}

\section{Additional Methodological Details}
\subsection{ODE-to-SDE for Exploration}
A fundamental challenge in applying GRPO to rectified flow models is that their standard sampling process is deterministic Ordinary Differential Equation (ODE)~\citep{ode}: $\mathrm{d}\mathbf{z}_t=\mathbf{u}_t\mathrm{d}t$. This deterministic nature prevents the stochastic exploration required for the policy to discover multiple trajectory. The key insight from prior work\citep{dancegrpo, flowgrpo} is to convert this deterministic ODE into a Stochastic Differential Equation (SDE) during training. This is achieved by introducing a controlled noise term into the reverse-time generative process:
\begin{equation}
    d\mathbf{z}_t = \left(u_\theta(\mathbf{z}_t, t) - \frac{1}{2}\varepsilon_t^2\nabla\log p_t(\mathbf{z}_t)\right)dt + \varepsilon_t d\mathbf{w}
    \label{sde:rf}
\end{equation}
where $\mathrm{d} \mathbf{w}$ is a Brownian motion, and $\varepsilon_t$ introduces the stochasticity during sampling. The SDE specifically designed to preserve the marginal distributions $p_t(\mathbf{z}_t)$ of the original ODE, enabling valid stochastic exploration for GRPO.

\begin{figure}[htbp]
  \centering
  \setlength{\abovecaptionskip}{0.2cm}
  \setlength{\belowcaptionskip}{-0.4cm}
  \includegraphics[width=0.5\textwidth]{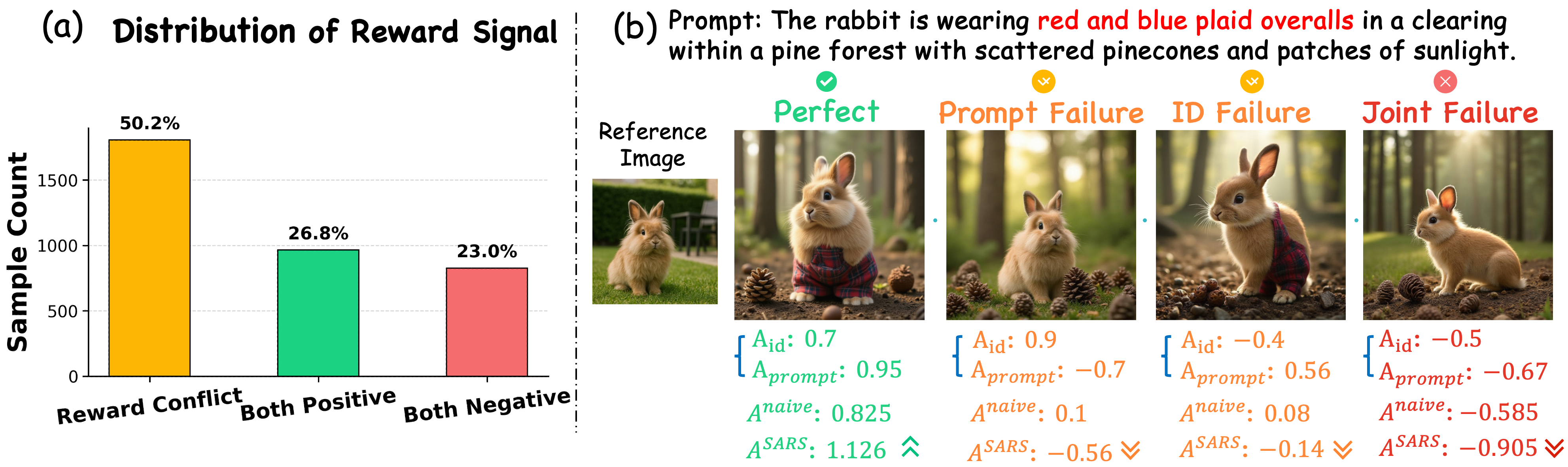}
  \caption{\textbf{The Impact of Reward Conflict and Our Synergy-Aware Reward Shaping}. (a) Distribution of advantage alignment, showing that reward conflict is the most common outcome, affecting 50.2\% of samples. (b) Comparison of final advantage calculation. Our method amplifies the signal for perfect outputs while correctly penalizing conflicted cases (Poor Prompt/ID) where Naive GRPO provides a weak, misleading signal.}
  \label{fig:conflict}
\end{figure}

\subsection{Detailed Analysis of Reward Conflict and SARS}
This section provides a detailed empirical analysis that motivates our Synergy-Aware Reward Shaping (SARS). As discussed in the main paper, a naive application of GRPO leads to Competitive Degradation. Our central hypothesis is that this macroscopic, policy-level failure stems from a microscopic, sample-level issue: a fundamental conflict in the reward signals themselves.

To validate this, we conducted a rigorous analysis of 300 training groups, each comprising 12 images, reveals a pervasive and fundamental issue we term \textbf{Reward Conflict}. 
As quantified in Figure~\ref{fig:conflict}, we find that within a typical group, approximately 50.2\% of the samples exhibit conflicting reward signals. 

This high incidence of conflict leads to two critical failures in the learning process. The linear aggregation causes the positive and negative advantages from these conflicted samples to neutralize each other. This consistently pushes the final advantage $A^{\text{naive}}_i$ towards zero, providing a weak and, more critically, misleading learning signal.

This misalignment with non-compensatory human preferences--where a single flaw is unacceptable--is evident in Figure~\ref{fig:conflict}(b). Naive GRPO assigns similarly low, positive advantages to a "Prompt Failure" ($A^{\text{naive}} =0.1$) and "ID Failure" ($A^{\text{naive}} = 0.08$), failing to distinguish between these critically flawed outputs. Consequently, the model receives an ambiguous gradient that incorrectly signals these "specialist" failures are acceptable, impeding effective policy optimization.

Our Synergy-Aware Reward Shaping (SARS) is explicitly designed to resolve this failure mode. As illustrated in the same case study in Figure~\ref{fig:conflict}(b). For a synergistic ("Perfect") output, SARS significantly amplifies the learning signal. More critically, it resolves the signal cancellation issue for conflicted samples. While Naive GRPO provides a weak and misleadingly positive signal for both "Prompt Failure" and "ID Failure" cases, SARS correctly identifies these as undesirable and transforms their advantage into a decisive negative penalty. 

\subsection{Customized-GRPO Algorithm}
We provide a detailed training procedure for our Customized-GRPO framework in Algorithm~\ref{algo:customized}. The overall process can be summarized as follows:
In each iteration, we first sample a batch of reference image and text prompt pairs from our training dataset. For each pair, we use the current policy $\pi_\theta$ to generate a group of $G$ candidate images. Next, we enter the core reward calculation phase. We compute the initial advantages for ID Preservation ( $A_{\text{id}}$ ) and Prompt Following ( $A_{\text{prompt}}$ ) for every generated image in the group. Following this, our two main contributions are applied:

\paragraph{Synergy-Aware Shaping:} We calculate the synergy term $\mathcal{S_i}$ for each sample based on the product of its advantages (Line 8).
\paragraph{Timestep-Aware Weighting:} We iterate through a subset of training timesteps. For each timestep $t$, we first retrieve the dynamic weights $(w_{\text{id}}(t), w_{\text{prompt}}(t))$ from our predefined function $f(t)$ (Line 12). These weights are then combined with the synergy term $\mathcal{S_i}$ to compute the final, time-dependent advantage $A_i^{\text{final}}(t)$ for each sample (Lines 14-18).

Finally, these time-dependent advantages are used to compute the GRPO objective $\mathcal{J}$.We then update the policy parameters $\theta$ via gradient ascent to maximize this objective (Line 21). This entire process is repeated until convergence.

\begin{table}[t]
    \centering
    \small
    \setlength{\abovecaptionskip}{0.2cm}
    \setlength{\belowcaptionskip}{-0.4cm}
    \resizebox{0.48\textwidth}{!}{
        \begin{tabular}{lcccc}
            \toprule
            \multirow{2}{*}[-0.2ex]{\textbf{Reward Model}} & \multicolumn{2}{c}{\textbf{ID Preservation}} & \multicolumn{2}{c}{\textbf{Prompt Following}}\\
            \cmidrule(lr){2-3} \cmidrule(lr){4-5}
            & \textbf{Pers.} & \textbf{Spear.} & \textbf{Pers.} & \textbf{Spear.} \\
            \midrule
            CLIP-I & 0.37 & 0.48 & - & -  \\
            DINO & 0.42 & 0.65 & - & -  \\
            CLIP-T & - & - & 0.38 & 0.51 \\
            \midrule
            Qwen-VL-Max & 0.27 & 0.42 & 0.41 & 0.57  \\
            GPT-4o-0816 & 0.38 & 0.63 & 0.52 & 0.69 \\
            Gemini-2.5-Pro & 0.49 & 0.68 & 0.57 & 0.72 \\
            \midrule
            \textbf{DINO-Seg} & \textbf{0.60} & \textbf{0.75} & - & - \\
            \textbf{HPS-V3} & - & - & \textbf{0.62} & \textbf{0.76}\\
            \bottomrule
        \end{tabular}
    }
    \caption{Comparisons of different reward model. Pers. and Spear. represents Person and Spearman correlations, respectively.}
    \label{tab: correlation}
\end{table}

\section{Additional Results}
\subsection{Impact of Reward Model Choice}
To further validate the robustness and generalizability of our Customized-GRPO framework, we investigate its performance with a diverse suite of reward models. Beyond our primary DINO-Seg and HPSv3 combination, we also experiment with CLIP-I and CLIP-T as alternative reward signals. Furthermore, inspired by recent work~\citep{t2i_factual, dreambench_plus, viescore, tifa, prism}, we evaluate the use of a powerful Vision-Language Model (VLM) including Gemini-2.5-Pro, Qwen-VL-Max and GPT-4o~\citep{gemini, qwenvlmax, GPT4o} as an automated evaluator for both ID preservation and prompt following.

\begin{table}[t]
    \centering
    \setlength{\abovecaptionskip}{0.2cm}
    \setlength{\belowcaptionskip}{-0.4cm}
    \resizebox{0.48\textwidth}{!}{
        \small
        \begin{tabular}{l|cccc}
            \toprule
            \textbf{Reward Model} & \textbf{DINO$\uparrow$} & \textbf{CLIP-I$\uparrow$} & \textbf{CLIP-T$\uparrow$} & \textbf{HPS-V3$\uparrow$}\\
            \midrule
            UNO (Base) & 0.760 & 0.835 & 0.304 & 7.87 \\
            \midrule
            DINO, CLIP-T & 0.731 & 0.782 & 0.287 & 7.12 \\
            CLIP-I, CLIP-T & 0.832 & 0.849 & 0.213 & 5.24 \\
            \midrule
            Qwen-VL-Max & 0.678 & 0.732 & 0.281 & 6.92 \\
            GPT-4o & 0.752 & 0.802 & 0.307 & 7.44 \\
            Gemini-2.5-Pro & 0.803 & 0.842 & 0.315 & 8.17\\ 
            \midrule
            \textbf{DINO-Seg, HPS-v3} & 0.812 & 0.862 & 0.301 & 8.32 \\
            \bottomrule
        \end{tabular}
    }
    \caption{Performance on DreamBench after training with different reward model combinations.}
    \label{tab:reward_dreambench}
\end{table}

\begin{table*}[t]
    \centering
    \small
    \resizebox{\textwidth}{!}{
        \begin{tabular}{llcccc}
            \toprule
            \textbf{Method} & \textbf{Training Paradigm} & \textbf{Iteration Time (s)} & \textbf{Training Time (h)} & \textbf{Sampling Time (s)} & \textbf{Performance (DINO / HPS-v3)} \\
            \midrule
            UNO (Base) & SFT & 2.32 & 3 & 8.1 & 0.762 / 7.43 \\
            Naive-GRPO & Online RL & 291.28 & 80 & 8.1 & 0.801 / 7.56 \\
            \midrule
            \textbf{Customized-GRPO} & Online RL + SARS + TDW & 311.82 & 80 & 8.1 & \textbf{0.812 / 8.32} \\
            \bottomrule
        \end{tabular}
    }
    \caption{Wall-clock time analysis and computational efficiency comparison evaluated on 8 NVIDIA A100 GPUs.}
    \label{tab:wall_clock}
\end{table*}

\paragraph{Human Alignment of Different Reward Model.}
We first assess how well various automated reward models align with human annotators. To do this, we created a validation set of 200 generated images, covering a diverse range of subjects and prompts. For each image, three expert annotators rated both ID preservation and prompt following on a scale of 0 to 5. The final human score for each criterion is the average of these three ratings, ensuring robustness. For the same set of images, we calculated scores from each automated reward model, using carefully designed prompts for the Vision-Language Models (VLMs). We then measured the agreement between the models' scores and the averaged human scores using both Pearson and Spearman correlation coefficients.

The results present in Table~\ref{tab: correlation}. For ID preservation, our proposed DINO-Seg achieves the highest correlation with human annotators (Speraman = 0.75), significantly outperforming both standard vision models like DINO and the best-performing VLM, Gemini-2.5-Pro. For prompt following, HPS-V3 demonstrates the strongest human alignment (Spearman = 0.76). Therefore, we choose DINO-Seg and HPS-V3 as reward models for Customized-GRPO training.
\paragraph{Performance with Different Reward Model Combinations.}
Following the alignment analysis, we trained Customized-GRPO using various combinations of reward models and evaluated their final performance on DreamBench. The results, detailed in Table~\ref{tab:reward_dreambench}, reveal several key insights.

First, the choice of reward model is critical to the final generation quality. Using reward models with lower human alignment, such as the CLIP-I, CLIP-T combination, leads to a significant degradation in performance, particularly in Prompt Following (HPS-V3 score drops to 5.24). This highlights that a reward signal misaligned with human preference can actively harm the policy optimization process.

Second, while powerful VLMs like Gemini-2.5-Pro can serve as effective reward models and yield strong results (HPS-V3 of 8.17), they do not surpass our specialized combination. We hypothesize this is because VLMs, while excellent at understanding high-level semantics, may be less sensitive to the fine-grained details crucial for ID Preservation. For instance, they might overlook subtle differences in texture, color shades, or small patches that are easily discernible to humans.

Ultimately, our primary combination of DINO-Seg and HPS-v3, which exhibited the highest human alignment in the previous analysis, also yields the best-balanced performance on the final benchmark.

\subsection{Computational Efficiency Analysis}
To comprehensively evaluate the computational efficiency of our proposed method, we conduct a wall-clock time analysis using 8 NVIDIA A100 GPUs on the 10k subset of the Syncd dataset. The comparison setup is configured as follows:
\begin{itemize}
    \item \textbf{SFT}: Batch size = 8, trained for 2 epochs.
    \item \textbf{GRPO (Naive \& Ours)}: Batch size = 1, Group size = 12, trained for 1 epoch.
\end{itemize}

The quantitative results regarding training and inference efficiency are summarized in Table~\ref{tab:wall_clock}. Based on these results, we derive two main observations:

\textbf{Training Efficiency:} While RL-based training ($\sim$80h) involves higher computational costs than standard SFT (3h) due to the inherent nature of online exploration and reward evaluation, it yields substantial performance gains across both objectives (DINO score increases from $0.762$ to $0.812$; HPS-v3 increases from $7.43$ to $8.32$). Crucially, compared to the Naive-GRPO baseline, our Customized-GRPO incurs \textbf{negligible additional overhead}. This demonstrates that our core contributions—Synergy-Aware Reward Shaping (SARS) and Time-Aware Dynamic Weighting (TDW)—are computationally lightweight and do not burden the overall RL training process.

\textbf{Sampling Efficiency:} Our method maintains a sampling latency ($8.1s$) identical to the base UNO model and Naive-GRPO. Since our optimization strictly updates model weights without altering the architecture, it introduces \textbf{zero additional cost during inference}.

\subsection{Qualitative Results}
\label{app:qualitative}
\begin{figure*}[htbp]
  \centering
  \includegraphics[width=1.0\textwidth]{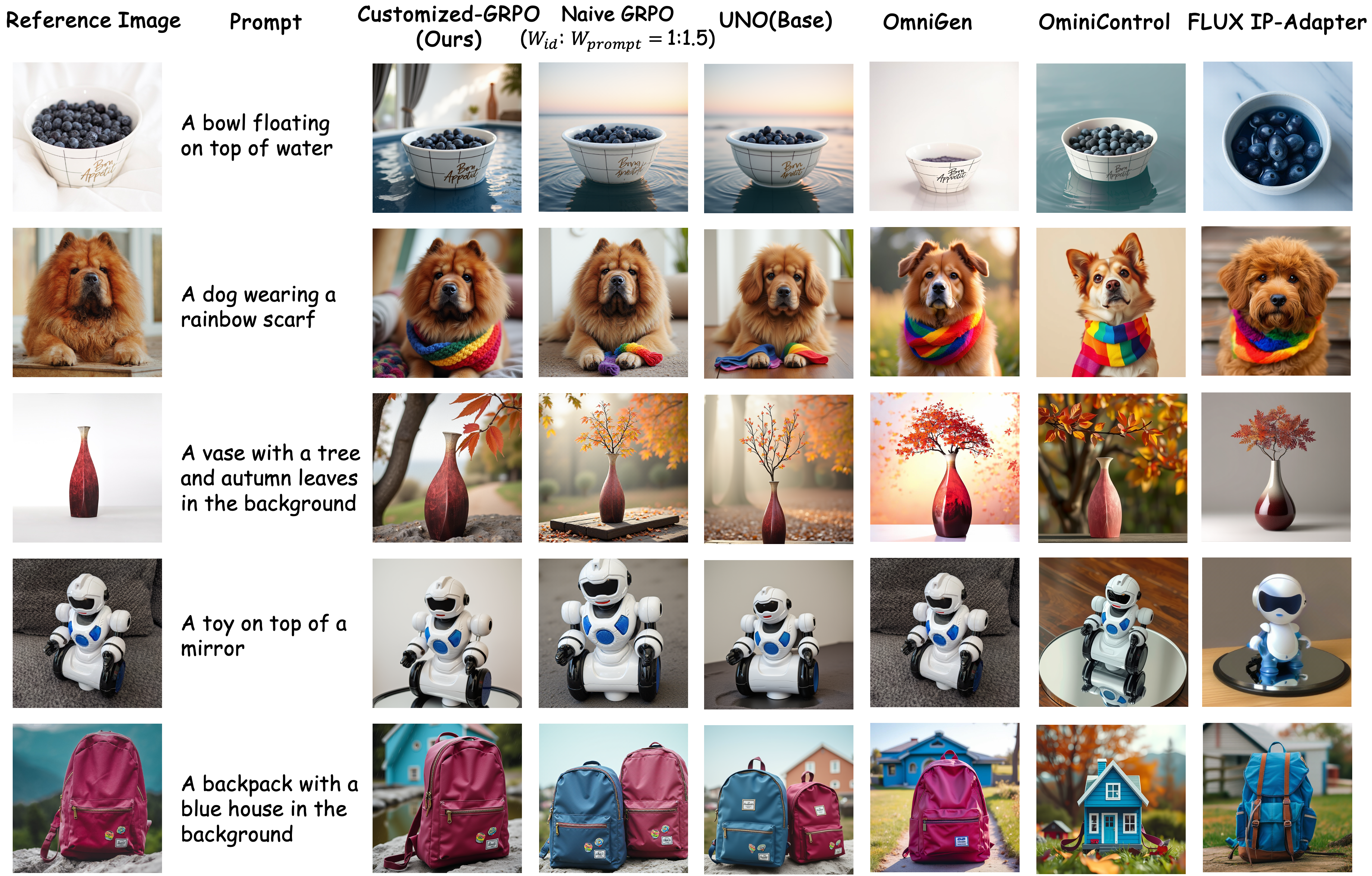}
  \caption{\textbf{Qualitative comparison of Customized-GRPO with state-of-the-art baselines.} Our method consistently generates higher-quality images that better balance identity preservation and prompt following. }
  \label{fig:qualitative}
  \vspace{-3mm}
\end{figure*}

Figure~\ref{fig:qualitative} presents a qualitative comparison of our Customized-GRPO against several state-of-the-art methods to visually substantiate the effectiveness of our approach. The examples highlight our model's superior ability to improve both demands of ID Preservation and Prompt Following.
In the first two rows, our method demonstrates the best ID Preservation. For the "dog" subject, our method is the only one that accurately captures the specific facial structure and dense fur texture of the reference dog. Similarly, for the "bowl of blueberries," our model faithfully reproduces the text and the shape.

The following three rows showcase performance on more complex compositional prompts. Our method consistently excels at both ID Preservation and Prompt Following. For instance, in the "vase with a tree" example, our model correctly places the vase within an autumn scene, avoiding the common hallucination of generating a tree growing out of the vase itself. In the final row, it successfully renders both the subject backpack and the specified "blue house" in the background, instead of a blue backpack.
Furthermore, a consistent qualitative improvement is observed in the overall aesthetic quality of our generations. Beyond simply satisfying the core objectives, the outputs from Customized-GRPO tend to be more coherent, with better lighting and composition.

\begin{figure*}[htbp]
  \centering
  \setlength{\abovecaptionskip}{0.2cm}
  \setlength{\belowcaptionskip}{-0.4cm}
  \includegraphics[width=1.0\textwidth]{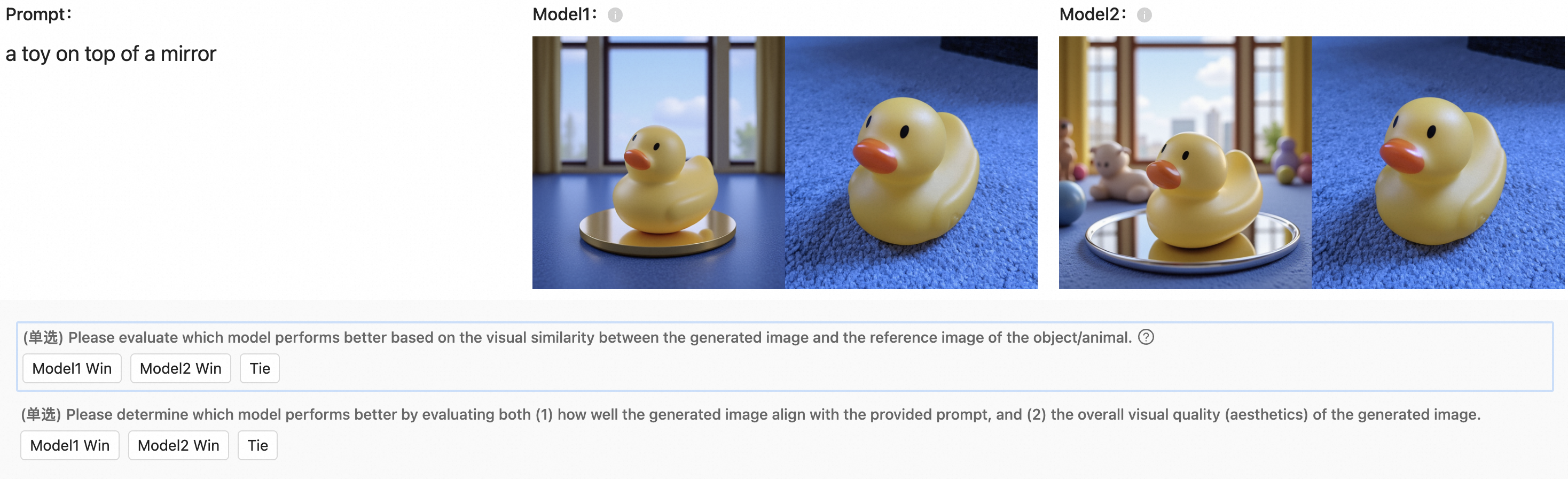}
  \caption{iTAG Interface for human evaluation.}
  \label{fig:user_interface}
  \vspace{-3mm}
\end{figure*}

\begin{figure*}[htbp]
  \centering
  \includegraphics[width=1.0\textwidth]{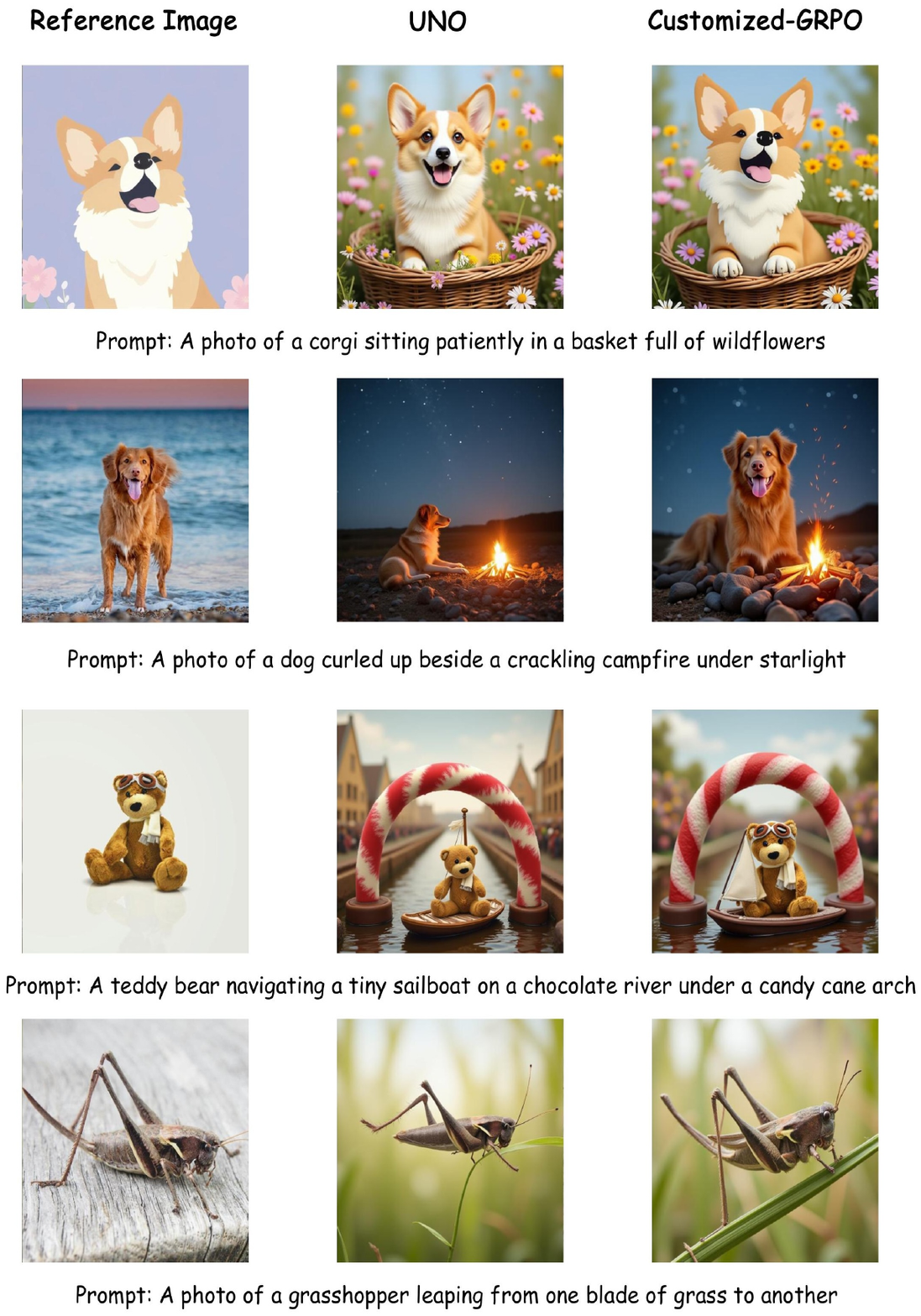}
  \caption{Comparison of the visualization results of UNO and Customized-GRPO}
  \label{fig:more_results_0}
\end{figure*}

\begin{figure*}[htbp]
  \centering
  \includegraphics[width=1.0\textwidth]{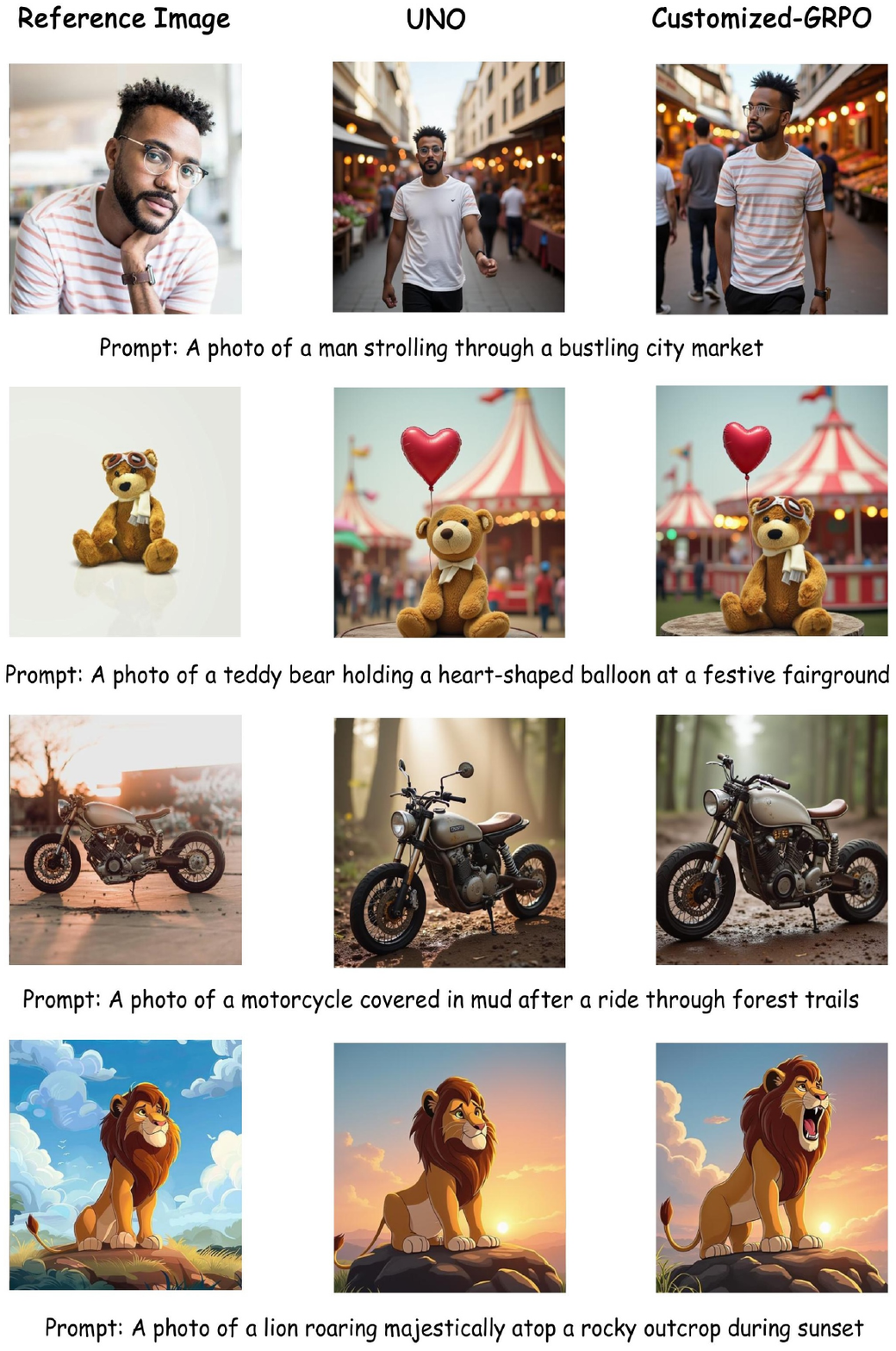}
  \caption{Comparison of the visualization results of UNO and Customized-GRPO}
  \label{fig:more_results_1}
\end{figure*}